\documentclass[10pt,journal,compsoc]{IEEEtran}

\ifCLASSOPTIONcompsoc
  \usepackage[nocompress]{cite}
\else
  \usepackage{cite}
\fi
\usepackage{times}
\usepackage{helvet}
\usepackage{courier}
\usepackage{amssymb}
\usepackage{amsmath}
\usepackage{amsfonts}
\usepackage{epsfig}
\usepackage{graphicx}
\usepackage{subfigure}
\usepackage{algorithmic}
\usepackage{algorithm}
\usepackage{multirow}
\usepackage{times}
\usepackage{xcolor}
\usepackage{soul}
\usepackage{mathrsfs}
\usepackage{amssymb}
\usepackage{graphicx}
\usepackage{psfrag}
\usepackage{subfigure}
\usepackage{stfloats}
\usepackage{float}
\usepackage{array}
\usepackage{amsthm}
\usepackage{booktabs}  
\usepackage{algorithm} 
\usepackage{algorithmic} 
\usepackage{amsmath,amsfonts,amssymb,amsbsy,bm,paralist,theorem,ifthen,color}

\ifCLASSINFOpdf
\else
\fi
\begin{document}

\title{GM-PLL: Graph Matching based Partial Label Learning}

\author{Gengyu~Lyu,
        Songhe Feng,
        Tao Wang,
        Congyan Lang,
        Yidong Li
\IEEEcompsocitemizethanks{\IEEEcompsocthanksitem School of Computer and Information Technology, Beijing Jiaotong University, No.3 Shangyuancun, Haidian District, Beijing 100044, China. \protect\\
E-mail: \{lvgengyu, shfeng, twang, cylang, ydli\}@bjtu.edu.cn}
}

\markboth{Journal of \LaTeX\ Class Files,~Vol.~14, No.~8, November~2018}%
{Shell \MakeLowercase{\textit{et al.}}: Bare Demo of IEEEtran.cls for Computer Society Journals}

\IEEEtitleabstractindextext{%
\begin{abstract}
Partial Label Learning (PLL) aims to learn from the data where each training example is associated with a set of candidate labels, among which only one is correct. The key to deal with such problem is to disambiguate the candidate label sets and obtain the correct assignments between instances and their candidate labels. In this paper, we interpret such assignments as instance-to-label matchings, and reformulate the task of PLL as a matching selection problem. To model such problem, we propose a novel \emph{Graph Matching based Partial Label Learning} (GM-PLL) framework, where Graph Matching (GM) scheme is incorporated owing to its excellent capability of exploiting the instance and label relationship. Meanwhile, since conventional \emph{one-to-one} GM algorithm does not satisfy the constraint of PLL problem that multiple instances may correspond to the same label, we extend a traditional \emph{one-to-one} probabilistic matching algorithm to the \emph{many-to-one} constraint, and make the proposed framework accommodate to the PLL problem. Moreover, we also propose a relaxed matching prediction model, which can improve the prediction accuracy via GM strategy. Extensive experiments on both artificial and real-world data sets demonstrate that the proposed method can achieve superior or comparable performance against the state-of-the-art methods.
\end{abstract}

}

\maketitle

\IEEEdisplaynontitleabstractindextext

\IEEEpeerreviewmaketitle

\IEEEraisesectionheading{\section{Introduction}\label{sec:introduction}}

\IEEEPARstart{A}{s} a weakly-supervised machine learning framework, partial label learning \footnote{In some literature, partial-label learning is also called as \emph{superset label learning} \cite{Liu:lotsllp-ICML2014},  \emph{ambiguous label learning}\cite{Chen:allud-IEEET2014} or \emph{soft label learning} \cite{Oukhellou2009Learning}.} learns from ambiguous labeling information where each training example corresponds to a candidate label set, among which only one is the ground-truth label \cite{wang2018towards} \cite{chen2013dictionary} \cite{zhang2014disambiguation}. During the training process, the correct label of each training example is concealed in its candidate label set and not directly accessible to the learning algorithm.

\begin{figure}
\centering
\setlength{\abovecaptionskip}{0.cm}
\setlength{\belowcaptionskip}{0.cm}
\includegraphics[width = 2.4in,height=1.2in]{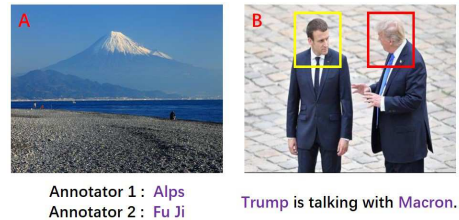}
\vspace{0mm}
\caption{Examplar applications of partial-label learning.}
\label{fig1}
\vspace{-5mm}
\end{figure}

In many real-world scenarios, data with explicit labeling information (unique and correct label) is too scarce to obtain than that with implicit labeling information (redundant labels). Thus, when faced with such ambiguous data, conventional supervised learning framework based on \emph{one instance one label} is out of its capability to learn from it accurately. Recently, Partial Label Learning (PLL) provides an effective solution to cope with it and has been widely used in many real-world scenarios. For example, in online annotation (Figure \ref{fig1} (A)), users with varying knowledge and cultural backgrounds tend to annotate the same image with different labels. In order to learn from such ambiguous annotated collection, it is necessary to find the correspondence between each image and its ground-truth label. In naming faces (Figure \ref{fig1} (B)), given a multi-figure image and its corresponding text description, the resulting set of images is ambiguously labeled if more than one name appear in the description. In other words, the specific correspondences between the faces and their names are unknown. In addition to the common scenarios mentioned above, PLL has also achieved competitive performance in many other applications, such as multimedia content analysis \cite{Zeng:lbaali-CVPR2013} \cite{xie2018pmll} \cite{Chen2018Learning} \cite{Cour:lfali-CVPR2009}, facial age estimation \cite{Zhang:pllvfad-TKDD2016}, web mining \cite{Luo:lfcls-NIPS2010}, ecoinformatics \cite{Liu:acmmmfsll-NIPS2012}, etc.

The key to accomplish the task of learning from Partial-Label (PL) data is disambiguation, which needs to fully explore the valuable information from ambiguous PL training data and obtain the correct assignments between the training INStances and their CandiDate Labels (INS-CDL). Recently, an Identification-based Disambiguation Strategy (IDS) is widely used in many PLL framework owing to its competitive performance on alleviating the interference of false positive labels \cite{Liu:acmmmfsll-NIPS2012} \cite{Jin:lwml-NIPS2003} \cite{feng2018leveraging} \cite{Nguyen:cwpl-KDDM2008} \cite{Yu:mmpll-ML2015} \cite{Zhang:stpllpaiba-IJCAI2015}. Among existing PLL methods based on IDS, some are often combined with the off-of-shelf learning schemes to identify the ground-truth label in an iterative manner, such as maximum likelihood \cite{Liu:acmmmfsll-NIPS2012} \cite{Jin:lwml-NIPS2003} \cite{feng2018leveraging}, maximum margin \cite{Nguyen:cwpl-KDDM2008} \cite{Yu:mmpll-ML2015} \cite{yu2016maximum}, etc. Others often try to explore the instance relationship from the ambiguous training data and directly disambiguate the candidate label sets \cite{Zhang:stpllpaiba-IJCAI2015}. Although the two kinds of PLL methods have obtained desirable performance in many real-world scenarios, they still suffer from some common defects. For example, for the instance relationship, they only consider the $k$-nearest-neighbor instances' similarity while simultaneously ignore the similarity among other instances and the dissimilarity among all instances, which makes the modeling output from unseen instance be overwhelmed by those from the negative nearest instances. And for the instance-label assignments, they usually utilize an iterative propagation procedure to implicitly obtain the objective labels, but neither explicitly describe the existing INS-CDL assignments relationship nor take the co-occurrence possibility of varying instance-label assignments into consideration to directly identify the optimal assignments, which may make the algorithm lose sight of direct instance-label assignments and result in its excessive attention to the instance relationship.

\begin{figure}
\centering
\setlength{\abovecaptionskip}{0.cm}
\setlength{\belowcaptionskip}{0.cm}
\includegraphics[width = 3in,height=1.5in]{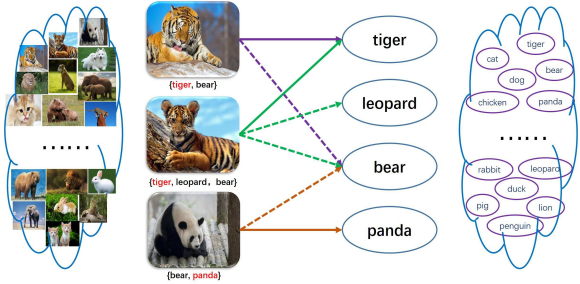}
\vspace{-1mm}
\caption{Illustration of formulating PLL as a matching selection problem.}
\label{fig2}
\vspace{-5mm}
\end{figure}
In order to overcome the above shortcomings, in this paper, we reinterpret the task of PLL as a matching selection problem, and simultaneously incorporate the instance relationship and the co-occurrence possibility of varying instance-label assignments into the same framework, then provide a novel solution for PLL problem. Specifically, we regard the INS-CDL correspondences as the instance-label matchings, and the task of PLL can be further reformulated as an instance-label matching selection problem (Figure \ref{fig2}), i.e. identifying the correct matching relationship between INStances and their Ground-Truth Labels (INS-GTL). Afterwards, the goal of the PLL problem is transformed into how to solve the matching selection problem and obtain the optimal instance-label assignments. Recently, Graph Matching (GM) provides an effective solution for such problem, and owing to its excellent performance on utilizing structural information of training data, it has been widely used in many real-world applications \cite{chertok2009spectral} \cite{egozi2010improving} \cite{hays2006discovering} \cite{wang2018gracker} \cite{wang2017graph}. Inspired by this, we incorporate the GM scheme into the PLL matching selection problem and propose a novel PLL learning framework named \emph{Graph Matching based Partial Label Learning} (GM-PLL). Note that, existing graph matching algorithms are formulated with \emph{one-to-one} constraint, which is not fully in accordance with the original task of PLL problem that one label can correspond to varying instances. Thus, we extend such \emph{one-to-one} constraint to \emph{many-to-one} constraint and propose a many-to-one probabilistic matching algorithm to make our method accommodate to the original PLL problem. Furthermore, during the establishment of the proposed framework, an affinity matrix is predetermined to describe the consistency relationship between varying INS-CDL assignments, where the similarity and dissimilarity of instances are simultaneously incorporated into the matrix. And these predetermined knowledge contributes the subsequent learning process and leads the algorithm to obtain the optimal solution. Moreover, to improve the predicted accuracy of test instances, we integrate the minimum error reconstruction scheme and graph matching scheme into a unified framework, and propose a relaxed GM predicted algorithm, where each unseen instance is first assigned with a candidate label set via minimum error reconstruction from its neighbor instances and then the predicted label is selected from $r$-maximum confidence candidate labels via graph matching strategy. Experimental results demonstrate that it can obtain higher classification accuracy than other predicted algorithms.

In summary, our main contributions lie in the following three aspects:
\begin{itemize}
\item Firstly, we reinterpret the conventional PLL problem and formulate the task of PLL as a matching selection problem. To the best of our knowledge, it is the first time to regard PLL problem as a matching selection problem, and accordingly we propose a novel GM-based PLL framework (GM-PLL), where instance relationship and the co-occurrence possibility of varying instance-label assignments are simultaneously taken into consideration.
\item Secondly, we extend conventional graph-matching algorithm with \emph{one-to-one} constraint to a probabilistic matching algorithm with \emph{many-to-one} constraint, which can guarantee that the proposed method fit the original task of PLL.
\item Finally, we propose a relaxed GM prediction algorithm, which simultaneously incorporate the graph matching scheme and minimum error reconstruction scheme into the same framework to improve the classification accuracy.
\end{itemize}

We start the rest of the paper by giving a brief introduction about PLL, and then present technical details of the proposed GM-PLL algorithm and the comparative experiments with existing state-of-the-art methods. Finally, we conduct experimental analysis and conclude the whole paper.

\section{Related Work}

Partial label learning, as a weakly supervised learning framework, focuses on solving the problem where data labeling information is excessively redundant. An intuitive strategy to cope with this issue is disambiguation, and existing disambiguation-based strategy are roughly grouped into three categories: \emph{Averaging Disambiguation Strategy} (ADS), \emph{Identification Disambiguation Strategy} (IDS) and \emph{Disambiguation-Free Strategy} (DFS). 

\subsection{Averaging Disambiguation Strategy (ADS)}
ADS-based methods usually assume that each candidate label has equal contribution to the learning model and they make prediction for unseen instances by averaging the outputs from all candidate labels. Following such strategy, Hullermeier et al. and Chen et al. adopt an instance-based model and disambiguate the ground-truth label by averaging the outputs of $k$-nearest neighbors following $\arg\max_{y\in\mathcal{Y}}\sum_{i\in\mathcal{N}_{({x}^{*})}}\!\!\mathbb{I}(y\!\!\in\!\!S_i)$ \cite{Huller:lfale-LNCS2005} \cite{Chen2017A}. Yu et al. utilize minimum error reconstruction criterion and obtain the predicted label via maximizing the confidence of $k$-nearest neighbors weighted-voting result \cite{Zhang:stpllpaiba-IJCAI2015}. Similarly, Tang et al. incorporate the boosting learning technique into its framework and improve the disambiguation classifier by adapting the weights of training examples and the ground-truth confidence of candidate labels \cite{tang:crdpll-AAAI2017}. Moreover, to further improve the disambiguation effectiveness, Zhang et al. facilitate its training process by taking the local topological information from feature space into consideration \cite{Zhang:pllvfad-TKDD2016}. Obviously, the above PLL methods are clear and easy to implement, but they share a critical shortcoming that the output of the ground-truth label is overwhelmed by the outputs of the other false positive labels, which will enforce negative influence on the disambiguation of ground-truth label.

\subsection{Identification Disambiguation Strategy (IDS)}
In order to overcome the shortcomings of ADS, the IDS based PLL methods are proposed to directly disambiguate the candidate label set. This strategy aims to build a direct mapping from instance space to label space, and accurately identify the ground-truth label for each training instance. Existing PLL algorithms following this strategy often view the ground-truth label as a latent variable first, identified as $\arg\max_{y\in{S_i}}F(\textbf{x},{\bm{\Theta}},y)$, and then refine the model parameter $\bm{\Theta}$ iteratively by utilizing Expectation-Maximization (EM) procedure \cite{Jin:lwml-NIPS2003}. Among these methods, some usually incorporate the maximum likelihood criterion and obtain the optimal label via maximizing the outputs of candidate labels, following $\sum_{i=1}^{n}\log (\sum_{y\in{S_i}}F(\textbf{x},{\bm{\Theta}},y))$ \cite{Chen:allud-IEEET2014} \cite{Liu:acmmmfsll-NIPS2012} \cite{Jin:lwml-NIPS2003} \cite{Grandvalet:lfplwme-CWP2004} \cite{Zhou2016Partial} \cite{vannoorenberghe2005partially}. Others often utilize the maximum margin criterion and identify the ground-truth label according to maximizing the margin between the outputs of candidate labels and that of the non-candidate labels, following $\sum_{i=1}^{n}(\max_{y\in{S_i}}F(\textbf{x},{\bm{\Theta}},y)-\max_{y\not\in{S_i}}F(\textbf{x},{\bm{\Theta}},y))$ \cite{Nguyen:cwpl-KDDM2008} \cite{Yu:mmpll-ML2015}. Experimental results demonstrate that IDS-based method has achieved superior and comparable performance than ADS-based methods.

\subsection{Disambiguation-Free Strategy (DFS)}
Recently, different from the two disambiguation-based PLL strategies mentioned above, some attempts have been made to learn from PL data by fitting the PL data to off-the-shelf learning techniques, where they can directly make prediction for the unseen instances without conduct the disambiguation on the candidate label set corresponding to the training instances. Following such strategy, Zhang et al. propose a disambiguation-free algorithm named PL-ECOC \cite{zhang:dfpll-IEEET2017}, which utilizes \emph{Error-Correcting Output Codes} (ECOC) coding matrix \cite{dietterich1994solving} and transfers the PLL problem into binary learning problem. Wu et al. propose another disambiguation-free algorithm called \emph{PALOC} \cite{TEBDfPLL-IJCAI2018}, which enables binary decomposition for PLL data in a more concise manner without relying on extra manipulations such as coding matrix. Experimental results empirically demonstrate that FDS-based algorithms can achieve comparable performance with the other disambiguation based PLL methods.

Although the above methods have achieved good performance on solving the PLL problem, they still suffer from some common shortcomings, i.e. they neither consider non $k$-nearest neighbor instance-similarity nor take the instance-dissimilarity into consideration. Therefore, in this paper, we utilize the GM scheme and propose a novel partial label learning framework called GM-PLL, where the instance similarity and dissimilarity are simultaneously incorporated into the framework to improve the performance of disambiguation. The details of the framework is introduced in the following section.


\section{The GM-PLL Method}
Formally speaking, we denote the $d$-dimensional input space as $\mathcal{X}\!=\!\mathbb{R}^{d}$, and the output space as $\mathcal{Y}\!=\! \{1,2,\ldots,\emph{q}\}$ with $q$ class labels. PLL aims to learn a classifier $f:\mathcal{X}\mapsto\mathcal{Y}$ from the PL training data $\mathcal{D}=\{(\textbf{x}_i,S_i)\} (1\leq{i}\leq{m})$, where the instance $\textbf{x}_{i}\in\mathcal{X}$ is described as a $d$-dimensional feature vector, the candidate label set $S_{i}=\{y_{i_1},y_{i_2},\ldots,y_{i_{|S_{i}|}}\}\subseteq\mathcal{Y}$ is associated with the instance $\textbf{x}_{i}$ and $|S_{i}|$ represents the number of candidate labels for instance $\textbf{x}_{i}$. Meanwhile, we denote $\textbf{y}=\{y_1,y_2,\ldots,y_m\}$ as the ground-truth label assignments for training instances, where each $y_{i}\in S_{i}$ corresponding to $\textbf{x}_{i}$ is not directly accessible to the algorithm.
\subsection{Formulation}
\begin{figure}[H]
\centering
\setlength{\abovecaptionskip}{0.cm}
\setlength{\belowcaptionskip}{0.cm}
\includegraphics[width = 2.2in,height=1.3in]{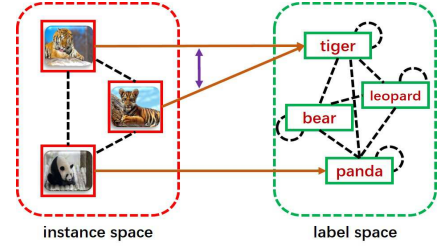}
\vspace{1mm}
\caption{The GM structure of GM-PLL. The GM structure originates from Figure \ref{fig2}. }
\label{fig3}
\end{figure}

GM-PLL is a novel PLL framework based on GM scheme, which aims to explore valuable information from ambiguous PL data and establish an accurate assignment relationship between the instance space $\mathcal{X}$ and the label space $\mathcal{Y}$. To make the proposed method easily understanding, we illustrate the GM-PLL method as a GM structure (Figure \ref{fig3}) before the following detailed introduction.

As depicted in Figure \ref{fig3}, both the instance space and label space are formulated as two different undirected graphs $\mathbb{G}^i=(\mathbb{V}^i,\mathbb{E}^i)$ of size $n_i$, where $i \in \{1, 2\}$, and $n_1 = m$, $n_2 = q$. The nodes $\mathbb{V}^i$ in the two graphs represent the instances and labels respectively, while the edges $\mathbb{E}^i$ encode their similarities. The goal of GM-PLL is to establish the graph nodes correspondence between $\mathbb{G}^1$ and $\mathbb{G}^2$.

Here, we first denote $\textbf{A}^{i}$ as the adjacent matrix for each graph $\mathbb{G}^i$, where $i=\{1,2\}$. $\textbf{A}^{1}\in\mathbb{R}^{m\times m}$ encodes the instance-similarity, which is calculated by normalizing the popular Cosine Metric,

\begin{flalign}
\textbf{A}^1_{ij} = \frac{\textbf{x}_{i}^\top \cdot \textbf{x}_{j}}{{||\textbf{x}_{i}||}_2\cdot{||\textbf{x}_{j}||}_2}
\label{Eq:ins-sim}
\end{flalign}
and $\textbf{A}^{2}\in\mathbb{R}^{q\times q}$ encodes the label-similarity,
\begin{flalign}
\textbf{A}^2_{i^{'}j^{'}} =
\begin{cases}
1, &\ \textrm{where} \ i^{'}= j^{'}\\
0, &\ \textrm{where} \ i^{'}\neq j^{'}\\
\end{cases}
\label{Eq:lab-sim}
\end{flalign}where the similarity of different labels is set to 0 owing to the inherent characteristics of PLL problem that the prior pairwise-label relationship is always missing. Note that once the label relationship as prior knowledge can be obtained, the proposed GM-PLL can still be easily extended to satisfy the problem.

Then, we define $\textbf{P}\in{\{0,1\}}^{m\times q}$ to describe the graph node correspondences between $\mathbb{G}^1$ and $\mathbb{G}^2$, where $\textbf{P}_{ij}=1$ represents label $j$ is assigned to instance $\textbf{x}_{i}$, and $\textbf{P}_{ij}=0$ otherwise. Among these correspondences that $\textbf{P}_{ij}=0$, a large number of them are invaluable to be considered since label $j$ is not contained in the candidate label set of instance $\textbf{x}_i$. Accordingly, we exclude the assignments between instances and their non-candidate labels, and obtain the row-wise vectorized replica $\textbf{p} = [p_1, p_2, \ldots, p_u]^\top\in \mathbb{R}^{u\times 1}$, where each element of $\textbf{p}$ is defined as:
\begin{flalign}
p_k = <\textbf{x}_{i_k},y_{l_k}>
\label{Eq:node-corres}
\end{flalign}
here $i_k\!\in\!\{1, 2, \ldots, m\}$, $l_k\!\in\!\{1,2,\ldots,|S_i|\}$, $u = \sum_{i=1}^{m}|S_i|$, $k\!\in\!\{1, 2, \ldots, u\}$ and the value of $<\!\!\textbf{x}_{i_k},y_{l_k}\!\!>$ represents the confidence of instance $\textbf{x}_{i_k}$ assigned with its $l_k$-th candidate label.

Afterwards, the correspondence of INS-CDL can be obtained by solving the optimization problem \textbf{OP (1)}
\begin{small}
\begin{align*}
\textbf{P}^{*} = \arg\max_{\textbf{P}}  \sum_{i_a,l_a,i_b,l_b} & d_{i_a,l_a,i_b,l_b}\textbf{P}_{i_a,l_a}\textbf{P}_{i_b,l_b}\\
s.t. \quad &\textbf{P}\textbf{\underline{1}}=\textbf{\underline{1}}.
\end{align*}
\end{small}where $d_{i_a,l_a,i_b,l_b}$ measures the pairwise consistency between instance edge $(i_a,i_b)$ and label edge $(l_a,l_b)$, which can also be regarded as the pairwise consistency between assignment $<\!\!\textbf{x}_{i_a}, y_{l_a}\!\!>$ and assignment $<\!\!\textbf{x}_{i_b},y_{l_b}\!\!>$. Motivated by recent studies \cite{wang2017graph} \cite{cho2010reweighted} \cite{liu2014gnccp}, we further formulate the \textbf{OP (1)} in a more general pairwise compatibility form \textbf{OP (2)}:
\begin{align*}
\textbf{p}^{*} = &\arg\max_{\textbf{p}} {\textbf{p}^{\top}\textbf{K} \textbf{p}}\\
s.t. \quad &\textbf{p}\in{\{0,1\}}^{u \times 1}\\
\quad &\textbf{P}\textbf{\underline{1}}= \textbf{\underline{1}}.
\end{align*}
where $\textbf{K}\in \mathbb{R}^{u\times u}$ is the affinity matrix that will be introduced in the following subsection \textbf{Generation of Affinity Matrix K}. And the optimization details of \textbf{OP (2)} will also be exhibited in the following Section \ref{section-optimization}.

\subsubsection{Generation of Affinity Matrix $\textbf{K}$}
Affinity Matrix $\textbf{K}\in \mathbb{R}^{u\times u}$ is defined to describe the matching consistency, and each element $\textbf{K}_{ab}$ represents the INS-CDL correspondence between $\textbf{p}_a$ and $\textbf{p}_b$, i.e.
\begin{flalign}
\textbf{K}_{ab} \ &= \ <\textbf{p}_{a},\textbf{p}_{b}> \nonumber \\
\ &=  \ <<\!\!\textbf{x}_{i_a},y_{{l}_a}\!\!>,<\!\!\textbf{x}_{i_b},y_{{l}_b}\!\!>>
\label{Eq:affinity-mean}
\end{flalign}
here $a,b \in \{1, 2, \ldots, u\}$, $<\!\!\textbf{x}_{i_a},y_{{l}_a}\!\!>$ represents the value of \emph{s}-th element of \textbf{p} as the INS-CDL correspondence between the \emph{$i_a$}-th instance $\textbf{x}_{i_a}$ and its \emph{$l_a$}-th candidate label $y_{l_a}$.

By predetermining the prior knowledge into the learning framework, affinity matrix can imply valuable information exploited from PL training data, including both the similarity and dissimilarity between instances, and the INS-CDL mapping relationship as well. Thus, we initialize the affinity matrix \textbf{K} as follows
\begin{flalign}
\textbf{K}_{ab} =
\centering
\begin{cases}
\textbf{A}^1_{ij},&\textbf{A}^2_{i^{'}j^{'}}=1\\
1 - \textbf{A}^1_{ij}, &\textbf{A}^2_{i^{'}j^{'}}=0.
\end{cases}
\label{Eq:affinity-value}
\end{flalign}

It is worth noting that, compared with the conventional PLL methods based on $k$-nearest neighbor scheme, the proposed framework contributes more prior knowledge to the learning process:
\begin{itemize}
\item[A)] It utilizes the similarity information from more training instances instead of only from the $k$-nearest neighbors.
\item[B)] It not only utilizes the instance similarity but also takes the dissimilarity between instances into consideration. Particularly, as shown in Eq (\ref{Eq:affinity-value}), with a higher similarity degree between two instances ($\textbf{x}_{i_a}$ and $\textbf{x}_{i_b}$), the $\textbf{K}_{ab}$ will get a higher value, i.e., the ground-truth labels ($y_{a}$ and $y_{b}$) of the two instances have higher probability to locate in the intersection of their candidate labels. On the contrary, if with a lower similarity degree between $\textbf{x}_{i_a}$ and $\textbf{x}_{i_b}$, $y_{a}$ and $y_{b}$ will have higher probability to belong to non-intersection of their candidate labels.
\end{itemize}

After initializing the affinity matrix $\textbf{K}$, we take the issue of class imbalance with respect to training data into consideration, and incorporate the number of instance candidate labels as a bias into the generation of affinity matrix:
\begin{small}
\begin{flalign}
\textbf{K}_{ab} = \textbf{K}_{ab} \cdot [1 + \alpha \cdot \mathrm{log}_{2}(\sum_{a = 1}^{u}h(\textbf{K}_{ab}>0)+\sum_{b = 1}^{u}h(\textbf{K}_{ab}>0))],
\label{Eq:affinity-balance}
\end{flalign}
\end{small}here $\alpha$ is the weight parameter, $h(\cdot)$ is the indicator function such that $h(\cdot) = 1$ iff $(\cdot)$ is true, and $h(\cdot) = 0$ otherwise. To reduce noise and alleviate the computational complexity, we increase the sparsity of the affinity matrix \textbf{K} and set $\textbf{K}_{ab} = 0$ if $\textbf{K}_{ab}<\beta$, where $\beta$ is the threshold parameter and it will be analyzed in Section \ref{section-analysis}.

At this point, the prior knowledge has been encoded into the affinity matrix, and it can provide good guidance for the subsequence learning process.

\begin{algorithm}[tb]
\caption{The Training Algorithm of \textbf{GM-PLL}}
\begin{algorithmic}
\label{Algorithm one}
   \STATE {\bfseries Inputs:}\\
   \quad $\mathcal{D}$: the partial label training set $\{(\textbf{x}_i,S_i)\}$;\\
   \STATE {\bfseries Process:}\\
   \STATE \textbf{1.} Calculate the cosine distances between each instance and derive the instance similarity matrix $\textbf{A}$ by Eq (\ref{Eq:ins-sim}); \\
   \STATE \textbf{2.} Calculate the affinity matrix $\textbf{K}$ by Eq (\ref{Eq:affinity-value}) and Eq (\ref{Eq:affinity-balance});\\
   \STATE \textbf{3.} Standardize the affinity matrix $\textbf{K}$ and remove low-confidence assignment by \textbf{{K(K$<\beta$) = 0}};\\
   \STATE \textbf{4.} Set $\textbf{K}^{(0)} = \textbf{K}$ and $\textbf{p}^{(0)}=\frac{1}{|S_i|}{\textbf{\underline{1}}}$ where $\textbf{p}^{(0)}\in{\mathbb{R}^{u\times 1}}$;\\
   \STATE \textbf{5.} \textbf{for} $t = 0$ \textbf{to} $iter$\\
   \STATE \textbf{6.}  \quad $\textbf{q}^{(t)} = \textbf{K}^{(t)}\textbf{p}^{(t)}$; \\
   \STATE \textbf{7.} \quad $\textbf{p}^{(t+1)}$ = Normalize($\textbf{q}^{(t)}$);\\
   \STATE \textbf{8.} \quad $\textbf{K}^{(t+1)}(a,b) = \textbf{K}^{(t)}(a,b)\cdot ({\textbf{p}^{(t+1)}_{a}}/{\textbf{p}^{(t)}_{a}})$;\\
   \STATE \textbf{9.} \quad \textbf{if} $({||\textbf{p}^{(t+1)}-\textbf{p}^{(t)}||}_2)<\delta$;\\
   \STATE \textbf{10.} \quad \quad break;\\
   \STATE \textbf{11.} \quad \textbf{end if}\\
   \STATE \textbf{12.} \textbf{end for}\\
   \STATE \textbf{13.} Discretize $\textbf{p}^{(t+1)}$, and derive the assignment ${(\textbf{x}_i,y_i)}$;\\
   \STATE {\bfseries Output:}\\
   \quad $y_i$: the assigned label for $\textbf{x}_i$;\\
\end{algorithmic}
\end{algorithm}

\subsection{Optimization}
\label{section-optimization}
In this section, we extend the probabilistic graph matching scheme from \cite{egozi2013probabilistic} and derive a probabilistic graph matching partial label learning algorithm. The core of the proposed algorithm is based on the observation that we can use the solution of the spectral matching algorithm \cite{Leordeanu2005A} to refine the estimate of the affinity matrix \textbf{K} and then solve a new assignment problem based on the refined matrix \textbf{K}. Namely, we can attenuate the affinities corresponding to matches with small matching probabilities and thus prune the affinity matrix \textbf{K}. In the same vein, we aim to adaptively increase the entries in \textbf{K} corresponding to assignments with high matching probabilities.

Concretely, we relax the first constraint of \textbf{OP (2)} to $\textbf{p}\in[0,1]^{u\times 1}$ and interpret \textbf{p} as matching probabilities $P(<\!\!\textbf{x}_i,y_{l}\!\!>)$. Then, the affinity matrix \textbf{K} can be further interpreted as a joint matching probabilities $P(<\!\!\textbf{x}_{i_a},y_{l_a}\!\!>,<\!\!\textbf{x}_{i_b},y_{l_b}\!\!>)$. Afterwards, we refine \textbf{K} and $\textbf{p}$ in an iterative manner where each iteration can be partitioned into two steps: estimating the mapping confidence of $\textbf{p}$ and refining the affinity matrix $\textbf{K}$. In the former step, we relax the \emph{one-to-one} constraints of \cite{Leordeanu2005A} as a \emph{many-to-one} constrain to accommodate that multiple instances may correspond to the same label. In the latter step, we follow \cite{egozi2013probabilistic} to make the refinement of $\textbf{K}$ allow analytic interpretation and provable convergence.

Hence, we minimize the objective function \textbf{OP (3)}

\begin{small}
\begin{align*}
[\textbf{p}^{*}_{a}, {(\textbf{p}_{a}|\textbf{p}_{b})}^*] = \arg\min_{a,b}\sum_{a}{((\sum_{b}(\textbf{p}_{a}|\textbf{p}_{b})\cdot \textbf{p}_{b})-\textbf{p}_{a})}^2
\end{align*}
\end{small}where $\textbf{p}_{a}$ is the assignment probability $P(<\!\!\textbf{x}_{i_a},y_{l_a}\!\!>)$ and $(\textbf{p}_{a}|\textbf{p}_{b})$ represents the conditional assignment probability $P(<\!\!\textbf{x}_{i_a},y_{l_a}\!\!>|<\!\!\textbf{x}_{i_b},y_{l_b}\!\!>)$ that is the probability of assignment $<\!\!\textbf{x}_{i_a},y_{l_a}\!\!>$ when $<\!\!\textbf{x}_{i_b},y_{l_b}\!\!>$ is valid. In our scheme, the $\textbf{p}_{a}$ and $(\textbf{p}_{a}|\textbf{p}_{b})$ need to be updated simultaneously.

Specifically, in iteration \emph{t}, we denote the estimation of $P^{(t)}(<\!\!\textbf{x}_{i_a},y_{l_a}\!\!>\!\!|\!\!<\!\!\textbf{x}_{i_b},y_{l_b}\!\!>)$ by $(\textbf{p}^{(t)}_{a}|\textbf{p}^{(t)}_{b})$ and $P^{(t)}(<\!\!\textbf{x}_{i_a},y_{l_a}\!\!>)$ by $\textbf{p}^{(t)}_{a}$, respectively. Then, we update $\textbf{p}^{(t)}_{a}$ by
\begin{flalign}
\textbf{p}^{(t+1)}_{a} =  \sum_{b}(\textbf{p}^{(t)}_{a},\textbf{p}^{(t)}_{b})= \sum_{b}(\textbf{p}^{(t)}_{a}|\textbf{p}^{(t)}_{b})\cdot \textbf{p}^{(t)}_{b}
\label{Eq:6}
\end{flalign}
where $(\textbf{p}^{(t)}_{a},\textbf{p}^{(t)}_{b})$ represents the joint probability $P(<\!\!\textbf{x}_{i_a},y_{l_a}\!\!>,<\!\!\textbf{x}_{i_b},y_{l_b}\!\!>)$ which is the joint probability of assignment $<\!\!\textbf{x}_{i_a},y_{l_a}\!\!>$ and assignment $<\!\!\textbf{x}_{i_b},y_{l_b}\!\!>$.

Different from the \emph{one-to-one} constraint of conventional GM problem, the framework of GM-PLL is formulated with \emph{many-to-one} constraint. Thus, we induce the constraint $\sum_{l_a=1}^{|S_i|}P(<\!\!\textbf{x}_{i_a},y_{l_a}\!\!>)$ =1. And $\textbf{p}^{(t+1)}_{a} = [\textbf{p}^{(t+1)}_{a_1}, \textbf{p}^{(t+1)}_{a_2}, \ldots, \textbf{p}^{(t+1)}_{a_{S_i}}]$ can be normalized as:
\begin{flalign}
\textbf{p}^{(t+1)}_{a_i} =  \frac{\textbf{p}^{(t+1)}_{a_i}}{\sum_{1}^{|S_i|}\textbf{p}^{(t+1)}_{a_i}}
\label{Eq:7}
\end{flalign}

Next, we refine the conditional assignment probability by
\begin{flalign}
{(\textbf{p}_{a}|\textbf{p}_{b})}^{(t+1)} =  {(\textbf{p}_{a}|\textbf{p}_{b})}^{(t)}\cdot \frac{{\textbf{p}_{a}}^{(t+1)}}{{\textbf{p}_{a}}^{(t)}}.
\label{Eq:8}
\end{flalign}

During the entire process of optimization, we first initialize the required variables, and then repeat the above steps until the algorithm converges. Finally, we get the assigned label for each training example. The whole training algorithm of GM-PLL is summarized in Algorithm \ref{Algorithm one}.

\subsection{Prediction}
\label{section-prediction}
During the stage of label prediction for unseen instances, we propose a graph matching based PLL prediction algorithm, which simultaneously takes the similarity reconstruction scheme and the GM scheme into consideration. The details of the prediction algorithm is introduced as follows.

We first integrate both the training instances and test instances into a large instances set, and then calculate a new instance-similarity matrix following Eq (\ref{Eq:ins-sim}). Afterwards, we assign the candidate label set for each test instance $\textbf{x}^{*}$ according to the weighted-voting results of its $k$-nearest neighbor instances $\mathcal{N}(\cdot)$, where the weights $\textbf{w}\in \mathbb{R}^{k\times 1}$ are calculated via minimum error reconstruction scheme \textbf{OP (4)}:
\begin{align*}
w_{c}^{*} = &\min_{w_{c}} {\bigg|\bigg|\textbf{x}^{*}-\sum_{c=1}^{k}w_{c}\cdot \textbf{x}_{c}\bigg|\bigg|}^{2} \\
s.t. \quad w_{c} \geq 0, \quad &\sum_{c=1}^{k} w_c = 1, \quad (\textbf{x}_{c} \in \mathcal{N}(\textbf{x}^{*}),1 \leq c \leq k)
\end{align*}here, $w_c$ is an element of $\textbf{w}$ and $c\in\{1, 2, \ldots, k\}$.

Based on the weighted-voting results, we obtain the confidence of each candidate label assigned to $\textbf{x}^{*}$, and then we can rank these labels according to the confidence in a descending order. Afterwards, we select the $r$-maximum confidence labels to constitute the candidate label set for $\textbf{x}^{*}$. Subsequently, the construction of candidate label set for each unseen instance has been completed.

Apparently, when the value of $r$ equals to the total number of candidate-label categories $q$, the predicted model will degenerate into disambiguation from all candidate labels, which is commonly in existing methods. In contrast, if only one label is retained ($r = 1$), the ground-truth label will be assigned with the maximum probability label, which is the same as \cite{Zhang:stpllpaiba-IJCAI2015}. The larger the value of $r$ is, the higher probability that the ground-truth label can be contained in the candidate label set, but meanwhile it would draw massive false labels that can decrease the effectiveness of the model. On the contrary, the smaller the value of $r$ is, the less false labels would be contained in the candidate label set, which would also result in the fact that the ground-truth label may be removed from the candidate label set.

Based on the above analysis, we can conclude that the total number of class labels (CL*) and the average number of class labels (AVG-CL*) for each instance have significant influence on the selecting of the number of assigned candidate labels $r$. Concretely, on one hand, more class labels means more noise class labels, thus we tend to assign $r$ with a smaller value to avoid the negative effect of these noise labels when CL* is larger. On the other hand, the average number of class labels can represent the average number of positive labels, thus we tend to choose larger $r$ when AVG-CL* is larger. At this point, we can calculate the $r$ by the following formula:
\begin{flalign}
r =  \bigg[1 + \frac{\textrm{AVG-CL*}}{{lg}(\textrm{CL*})}\bigg]
\label{induce_r}
\end{flalign}here $[\bigtriangleup]$ is the integral function, which represents the rounding operation for $\bigtriangleup$.

Finally, once the above operations are completed, we follow the idea of Algorithm \ref{Algorithm one} to rebuild the affinity matrix and utilize the GM scheme to recover the correct mapping between test instances and their ground-truth labels.


\section{Experiments}
\subsection{Experimental Setup}
To verify the effectiveness of the proposed GM-PLL method, we conduct experiments on \textbf{nine} controlled UCI data sets and \textbf{six} real-world data sets:

\textbf{(1) Controlled UCI data sets}. Under specified configuration of two controlling parameters (i.e. $\emph{p}$ and $\emph{r}$), the nine UCI data sets generate 189 ($7\times3\times9$) artificial partial-label data sets \cite{Chen:allud-IEEET2014} \cite{Cour:lfpl-JMLR2011}. Here, $\emph{p}\!\in\!{\{0.1,0.2,\ldots,0.7\}}$ is the proportion of instances with partial labeling and $\emph{r}\!\in\!{\{1,2,3\}}$ is the number of candidate labels except the ground-truth label. Table \ref{table1} summarizes the characteristics of the nine UCI data sets, including the number of examples (\textbf{EXP*}), the number of the features (\textbf{FEA*}), the whole number of class labels (\textbf{CL*}) and their common configurations (\textbf{CONFIGURATIONS}).

\begin{table}[!ht]
\centering
\caption{Characteristics of the controlled data sets}
\vspace{1mm}
\label{table1}
\small
\resizebox{9cm}{!}{
\begin{tabular}{cccc|c}
\hline \hline
UCI data sets  & EXP*     & FEA*     & CL*        &CONFIGURATIONS          \\ \hline
Glass         & 214      & 10       & 7           &          \\
Ecoli         & 336      & 7        & 8           &  \\
Dermatology   & 364      & 23       & 6           & $r = 1,  p\in\{0.1, 0.2, \ldots,0.7\}$  \\
Vehicle       & 846      & 18       & 4           & \\
Segment       & 2310     & 18       & 7           & $r = 2,  p\in\{0.1, 0.2, \ldots,0.7\}$\\
Abalone       & 4177     & 7        & 29          & \\
Letter        & 5000     & 16       & 26          & $r = 3,  p\in\{0.1, 0.2, \ldots,0.7\}$\\
Satimage      & 6345     & 36       & 7           &\\
Pendigits     & 10992    & 16       & 10          &\\ \hline \hline
\end{tabular}}
\vspace{-1mm}
\end{table}

\textbf{(2) Real-World (RW) data sets }. These data sets are collected from the four following task domains: (A) {\textbf{\emph{Facial Age Estimation}}} Human faces are represented as instances and the ages annotated by ten crowd-sourced labelers together with the ground-truth ages are regarded as candidate labels; (B) {\textbf{\emph{Automatic Face Naming}}} Human faces copped from images or videos are represented as instances and each candidate label set is composed of the names extracted from the corresponding captions or subtitles; (C) {\textbf{\emph{Object Classification}}} Image segmentations constitute the instance space and the objects appearing within the same image constitute the candidate label sets; (D) {\textbf{\emph{Bird Song Classification}}} Singing syllables of the birds are represented as instances while bird species jointly singing during a 10-seconds period are regarded as candidate labels; Table \ref{table2} summarizes the characteristics of the above real world data sets, including not only the number of examples (\textbf{EXP*}), the number of the feature (\textbf{FEA*}) and the whole number of class labels (\textbf{CL*}), but also the average number of class labels (\textbf{AVG-CL*}) and their task domains (\textbf{TASK DOMAIN}).

\begin{figure*}[!ht]
\centering
\begin{tabular}{ccc}
\includegraphics[width = 2in,height=1.6in]{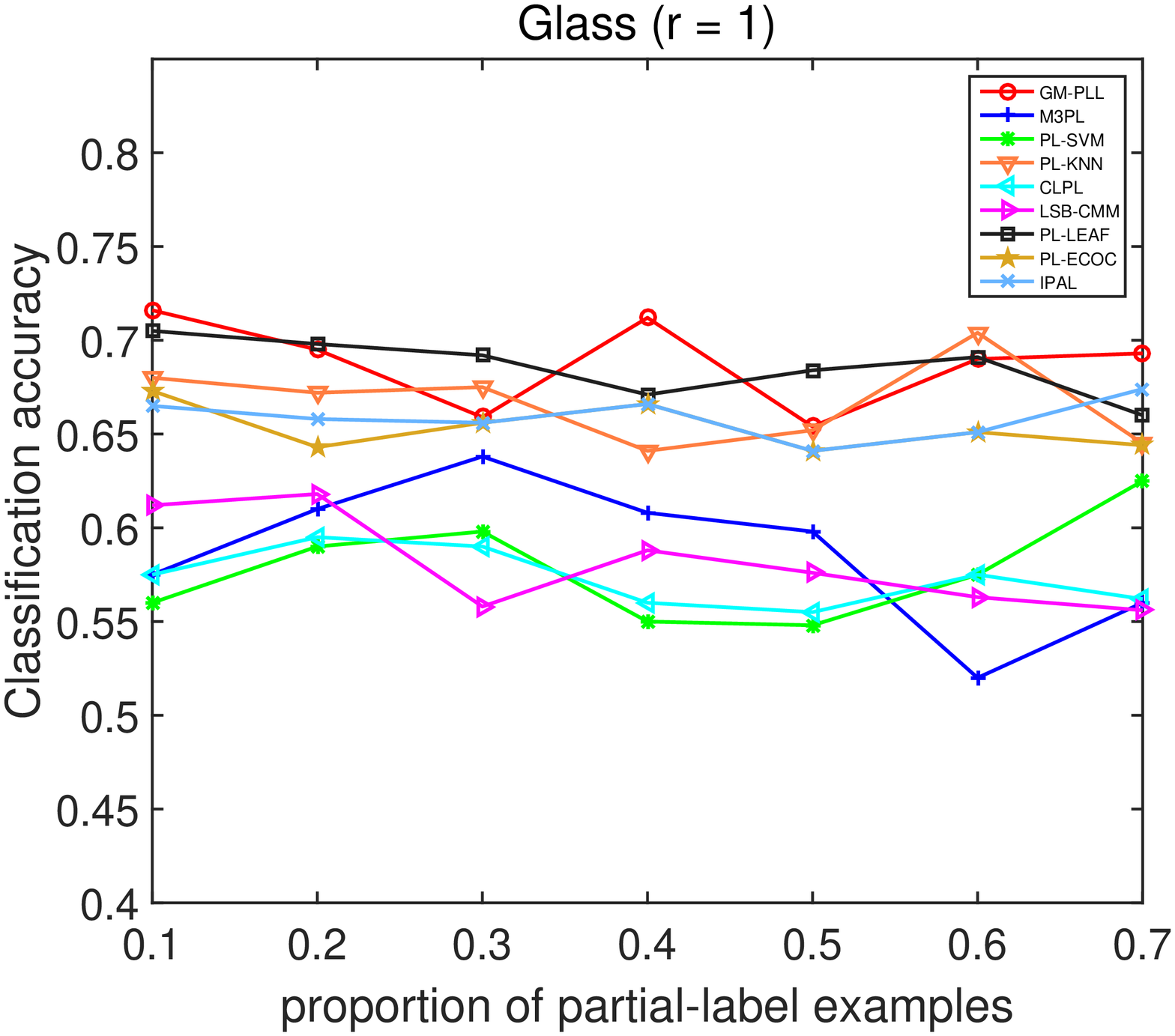}&\includegraphics[width = 2in,height=1.6in]{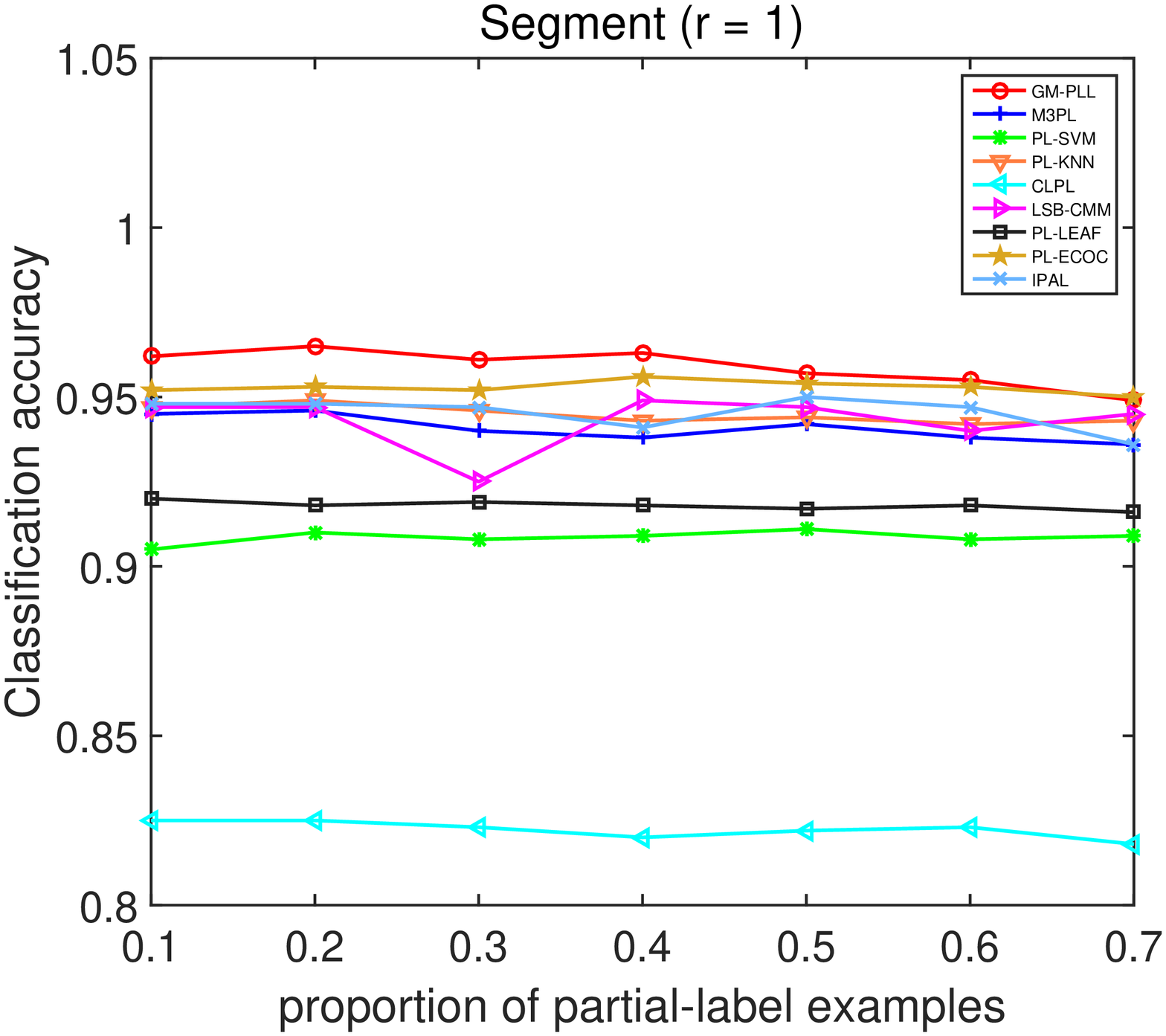}&\includegraphics[width = 2in,height=1.6in]{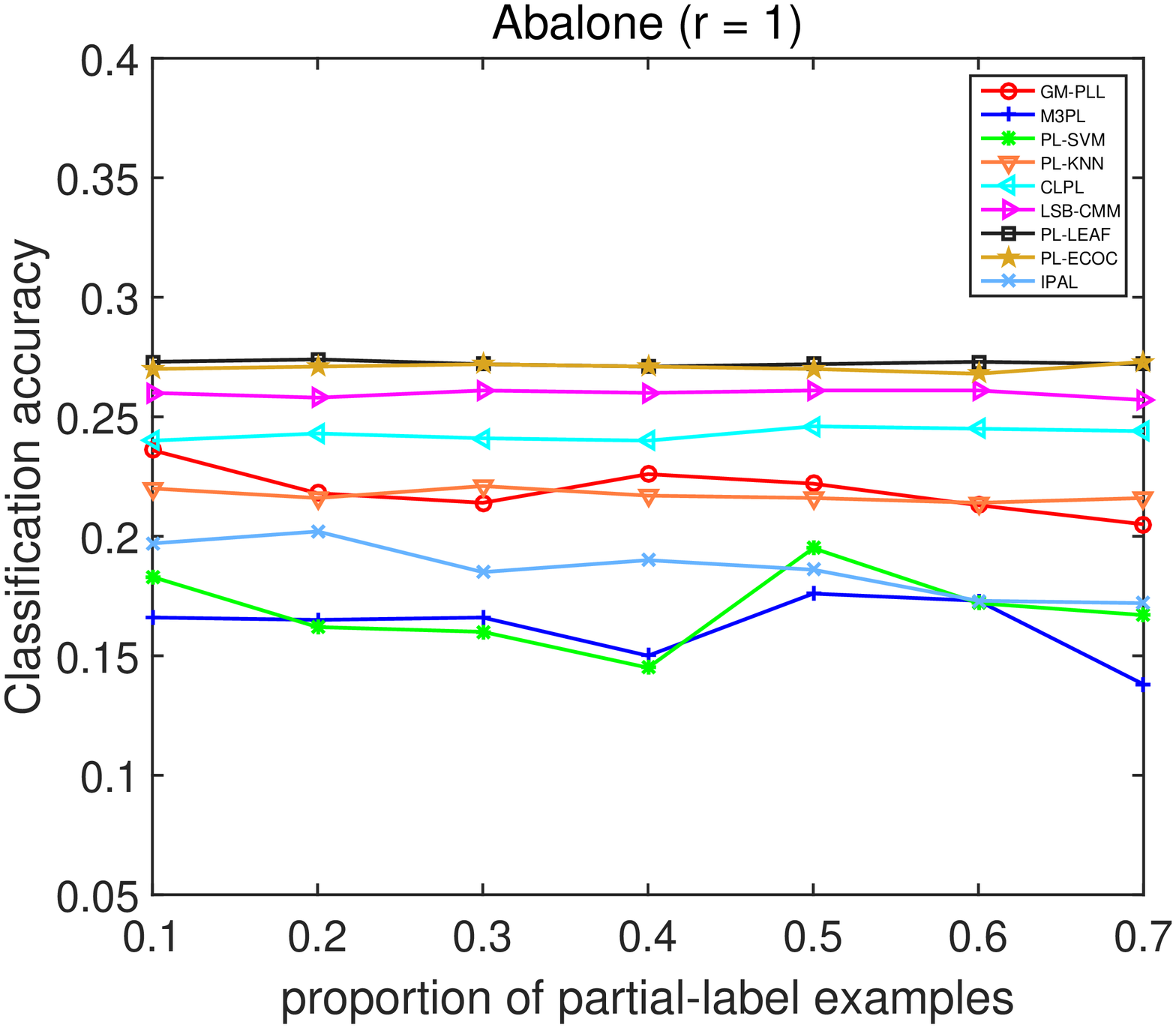}\\
\includegraphics[width = 2in,height=1.6in]{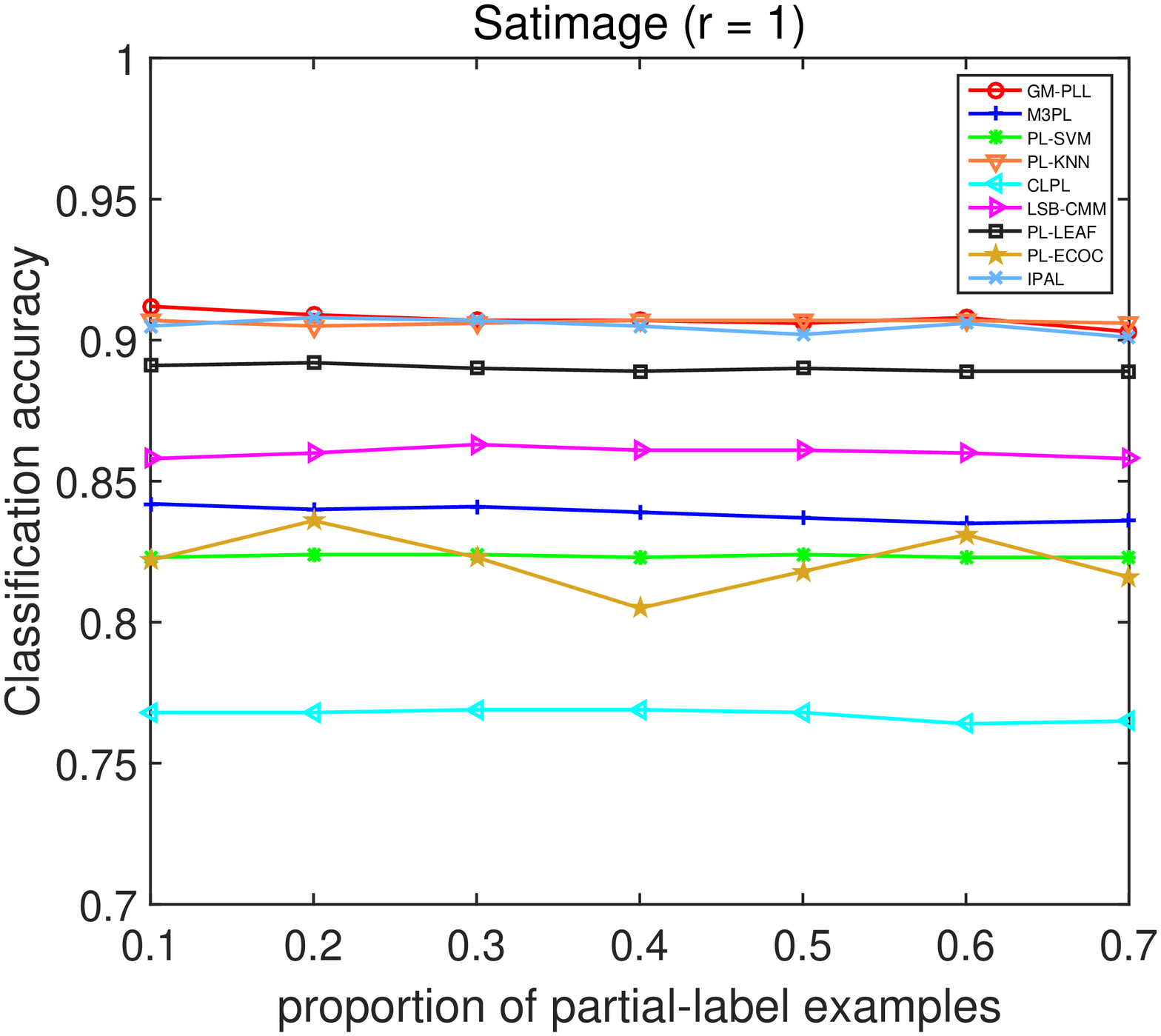}&\includegraphics[width = 2in,height=1.6in]{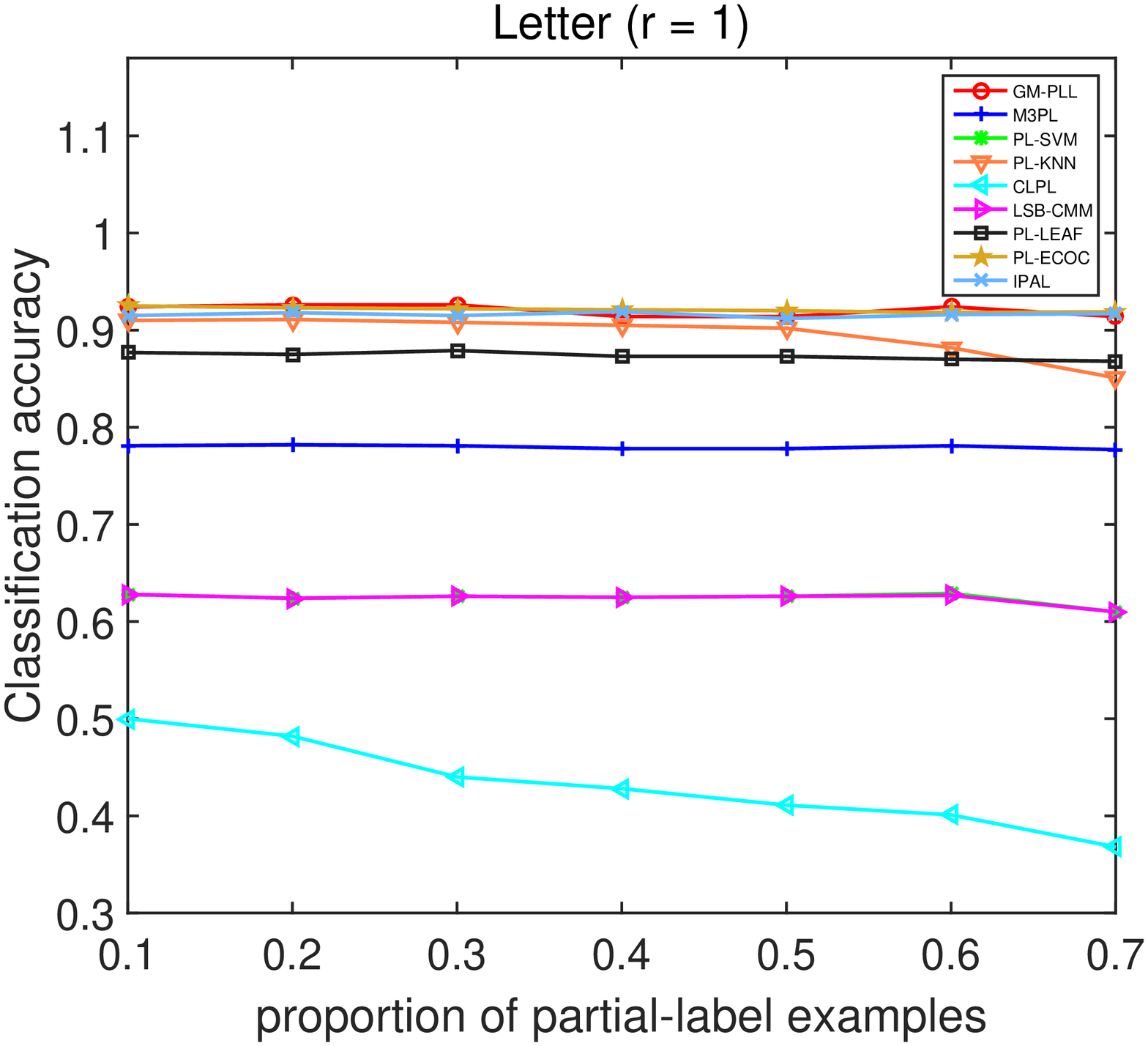}&\includegraphics[width = 2in,height=1.6in]{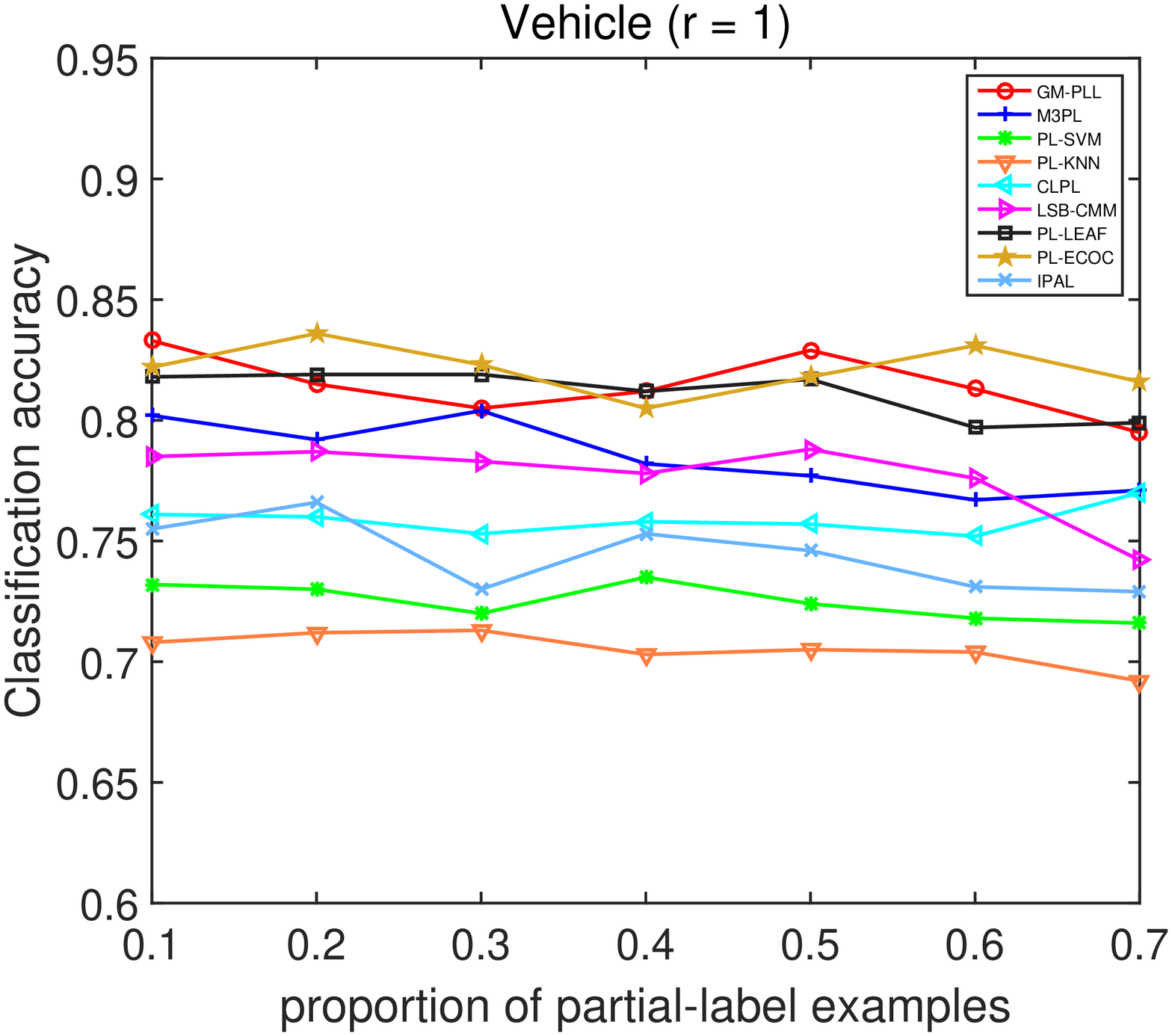}\\
\includegraphics[width = 2in,height=1.6in]{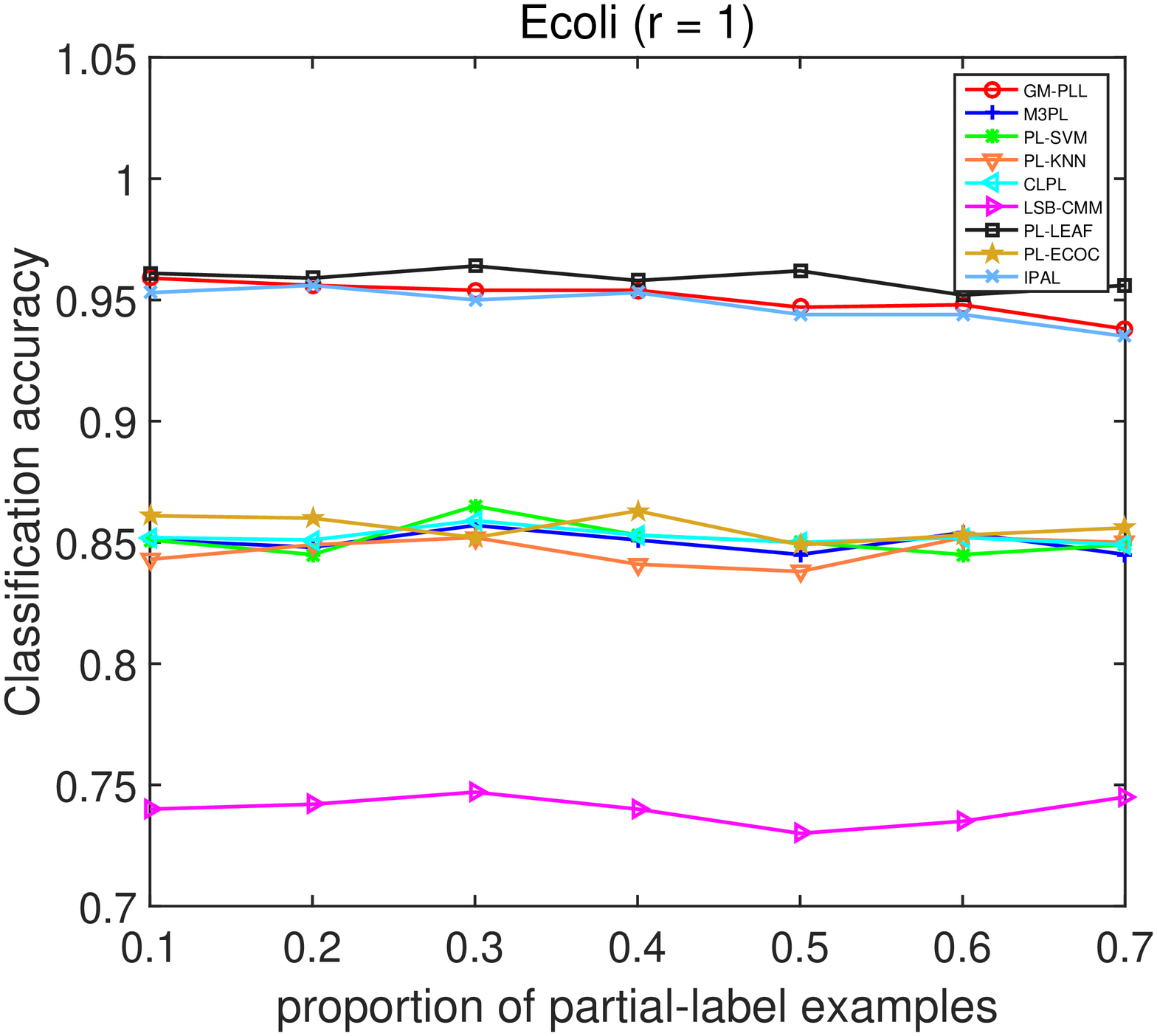}&\includegraphics[width = 2in,height=1.6in]{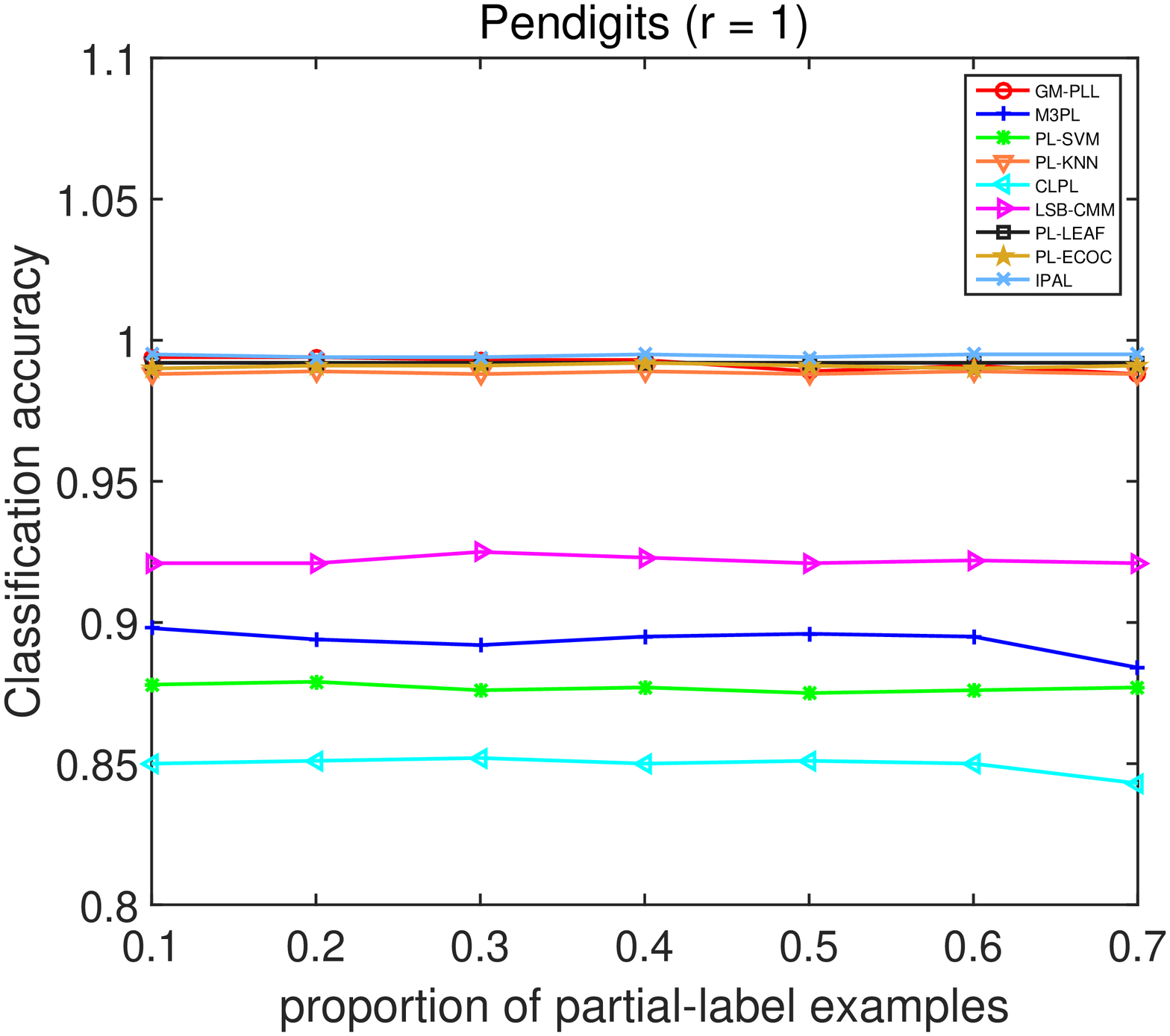}&\includegraphics[width = 2in,height=1.6in]{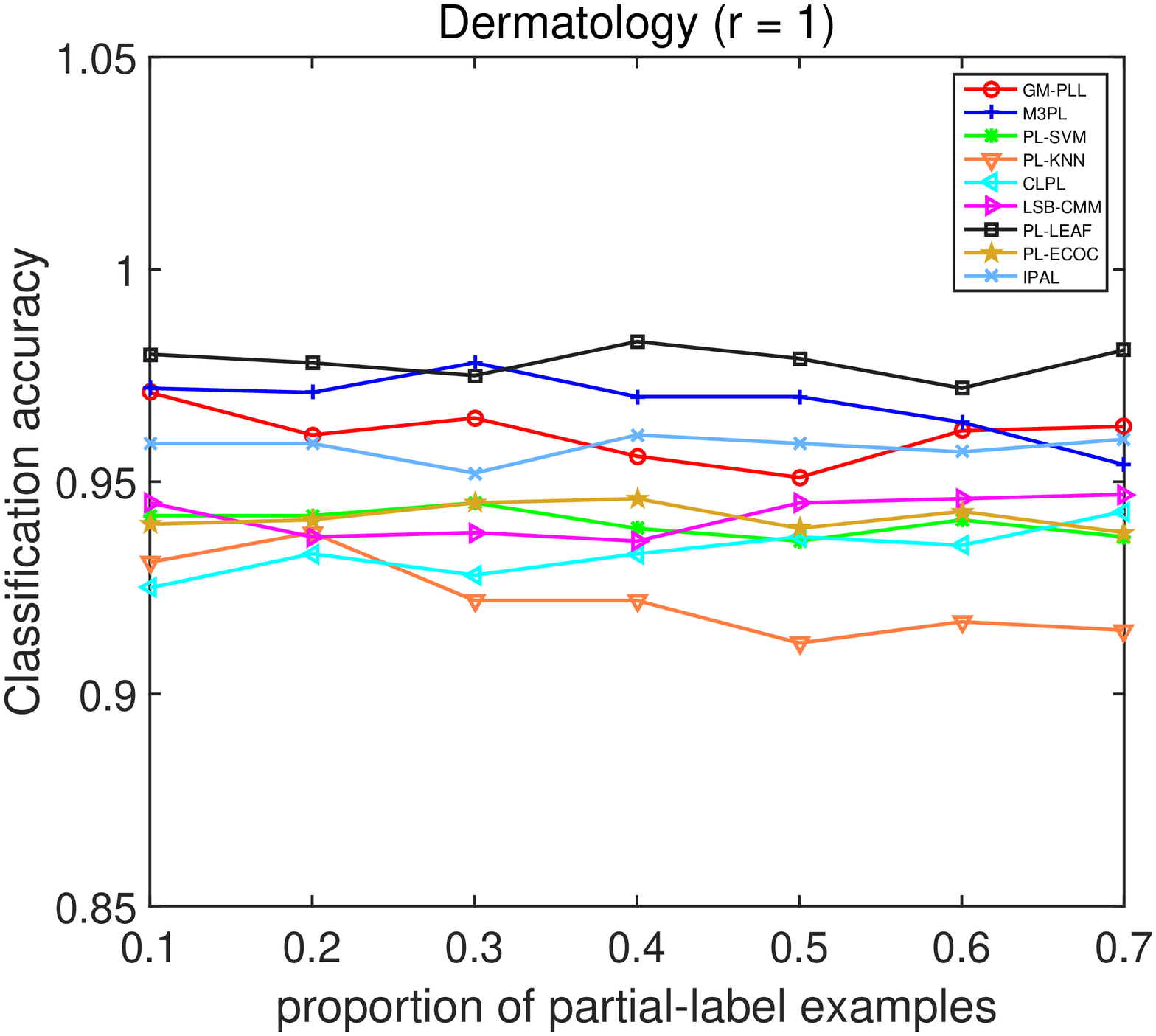}\\
\end{tabular}
\vspace{3mm}
\caption{The classification accuracy of each comparing method on nine controlled UCI data sets with one false positive candidate label (r = 1)}
\label{Fig-r1}
\vspace{0mm}
\end{figure*}

\begin{table*}[!ht]
\centering
\caption{Characteristics of the real-world data sets}
\vspace{1mm}
\label{table2}
\resizebox{13cm}{!}{
\begin{tabular}{cccccc}
\hline \hline
RW data sets   & EXP*     & FEA*     & CL*         & AVG-CL*   &TASK DOMAIN     \\ \hline

Lost          & 1122     & 108      & 16          & 2.33      &\emph{Automatic Face Naming} \cite{Cour:lfpl-JMLR2011}      \\
MSRCv2       & 1758     & 48       & 23          & 3.16      &\emph{Image Classification} \cite{Briggs:rlsimfmia-KDDM2012}  \\
FG-NET        & 1002     & 262      & 99          & 7.48      &\emph{Facial Age Estimation} \cite{Panis:FG-NET-JAH2015} \\
Soccer Player & 17472    & 279      & 171         & 2.09      &\emph{Automatic Face Naming} \cite{Zeng:lbaali-CVPR2013} \\
Yahoo! News & 22991    & 163      & 219         & 1.91      &\emph{Automatic Face Naming} \cite{Guill:mimlfalbof-ECCV2010} \\ \hline \hline
\end{tabular}}
\vspace{-1mm}
\end{table*}

\begin{figure*}[!ht]
\centering
\begin{tabular}{ccc}
\includegraphics[width = 2in,height=1.6in]{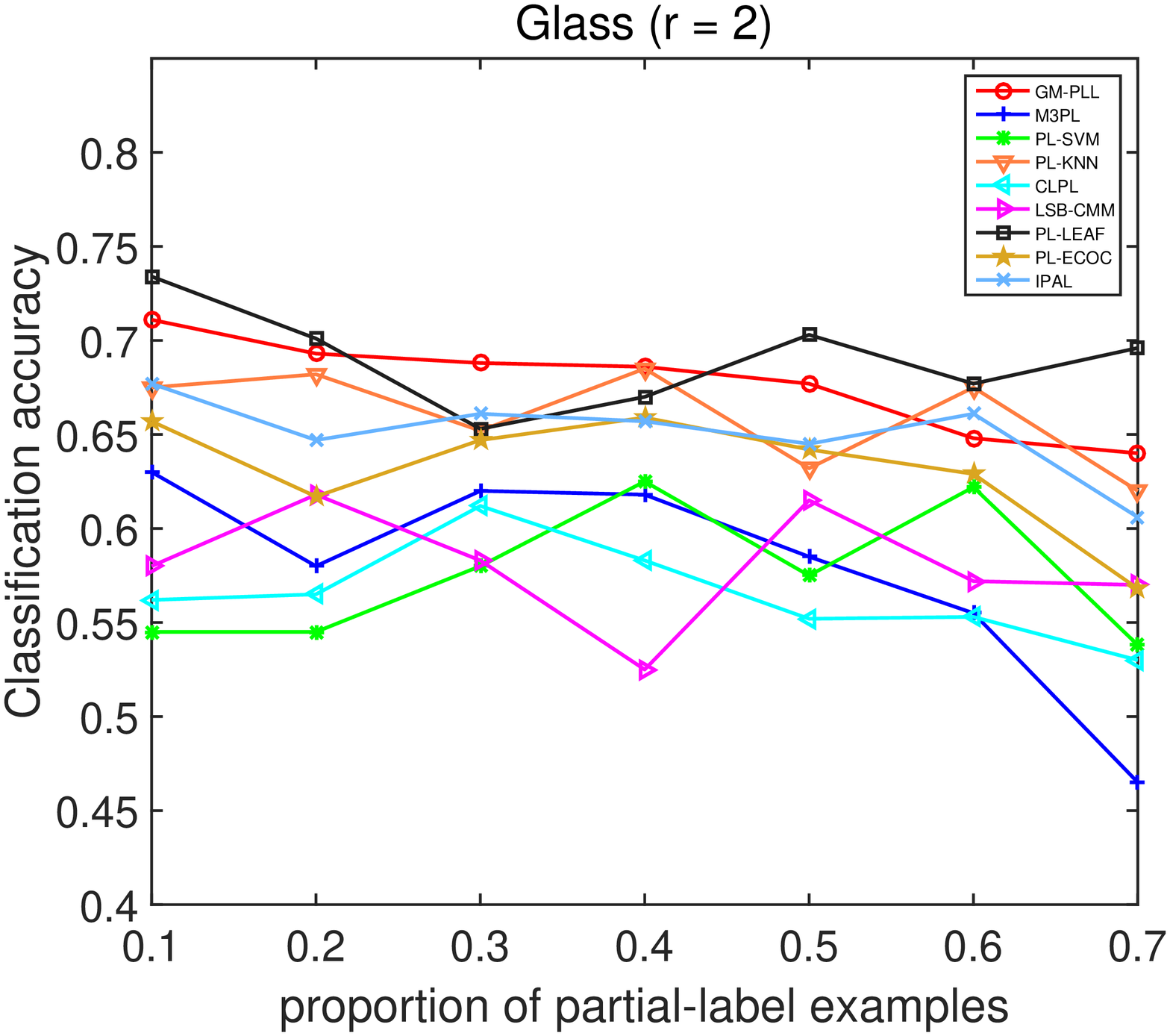}&\includegraphics[width = 2in,height=1.6in]{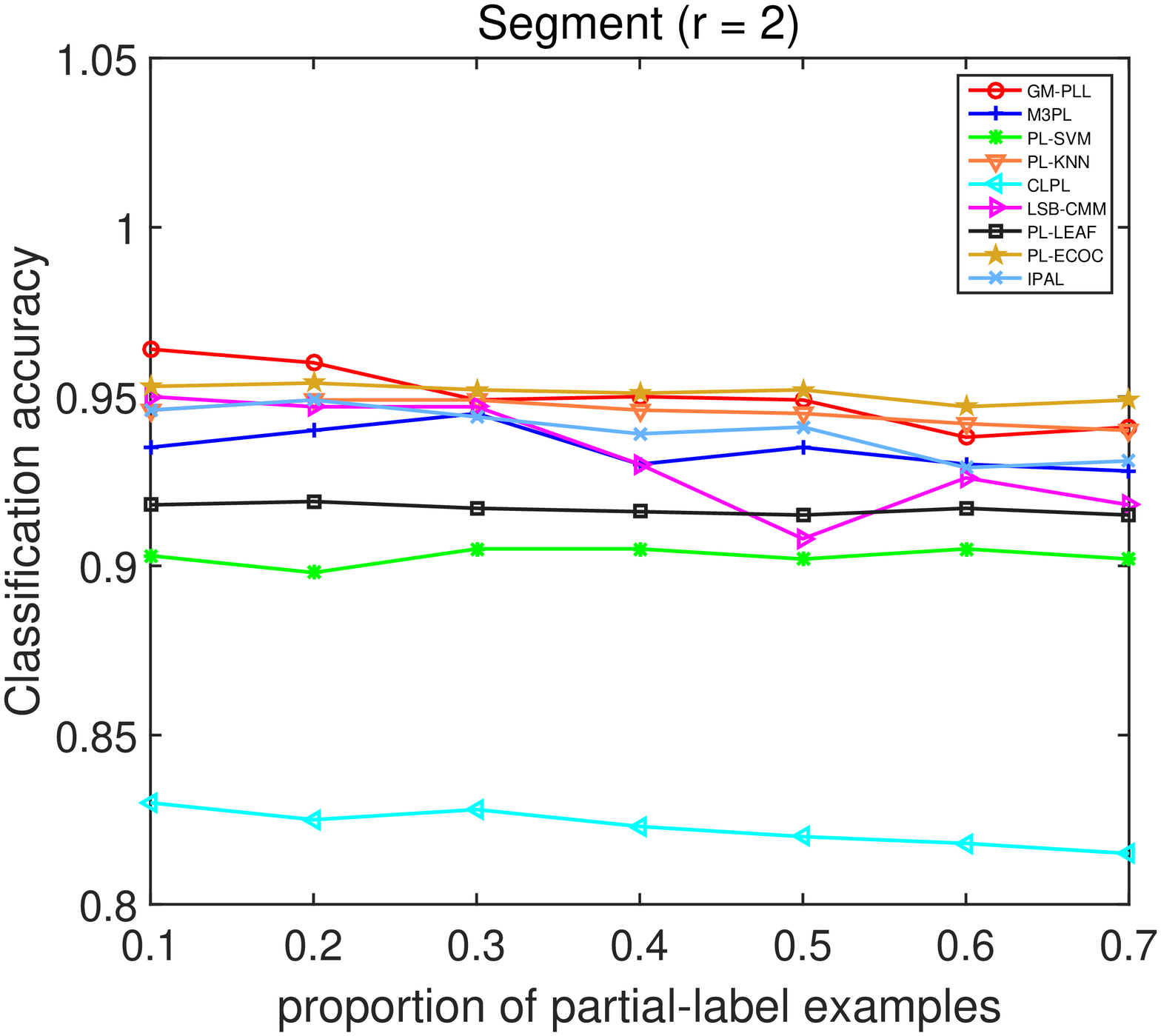}&\includegraphics[width = 2in,height=1.6in]{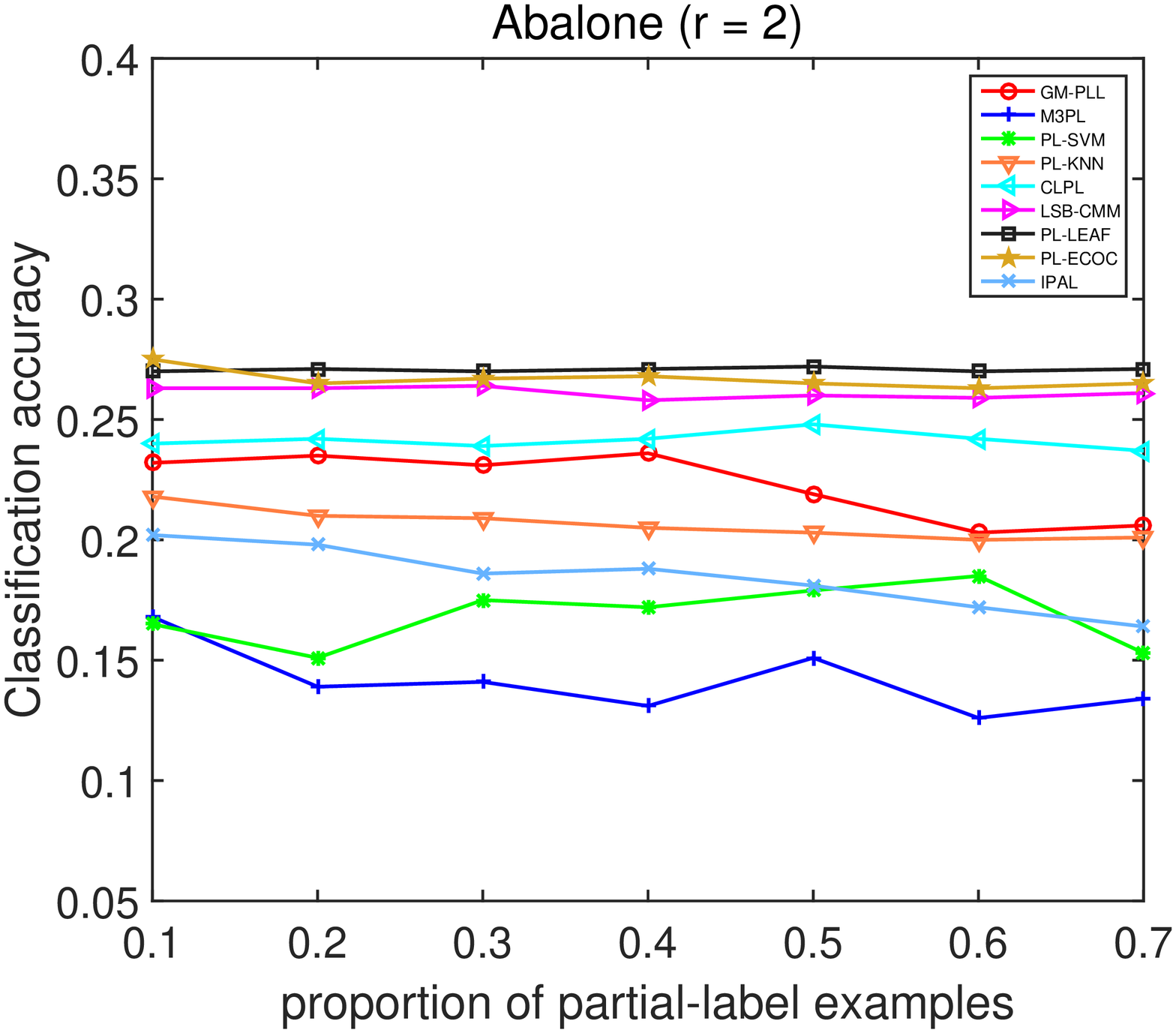}\\
\includegraphics[width = 2in,height=1.6in]{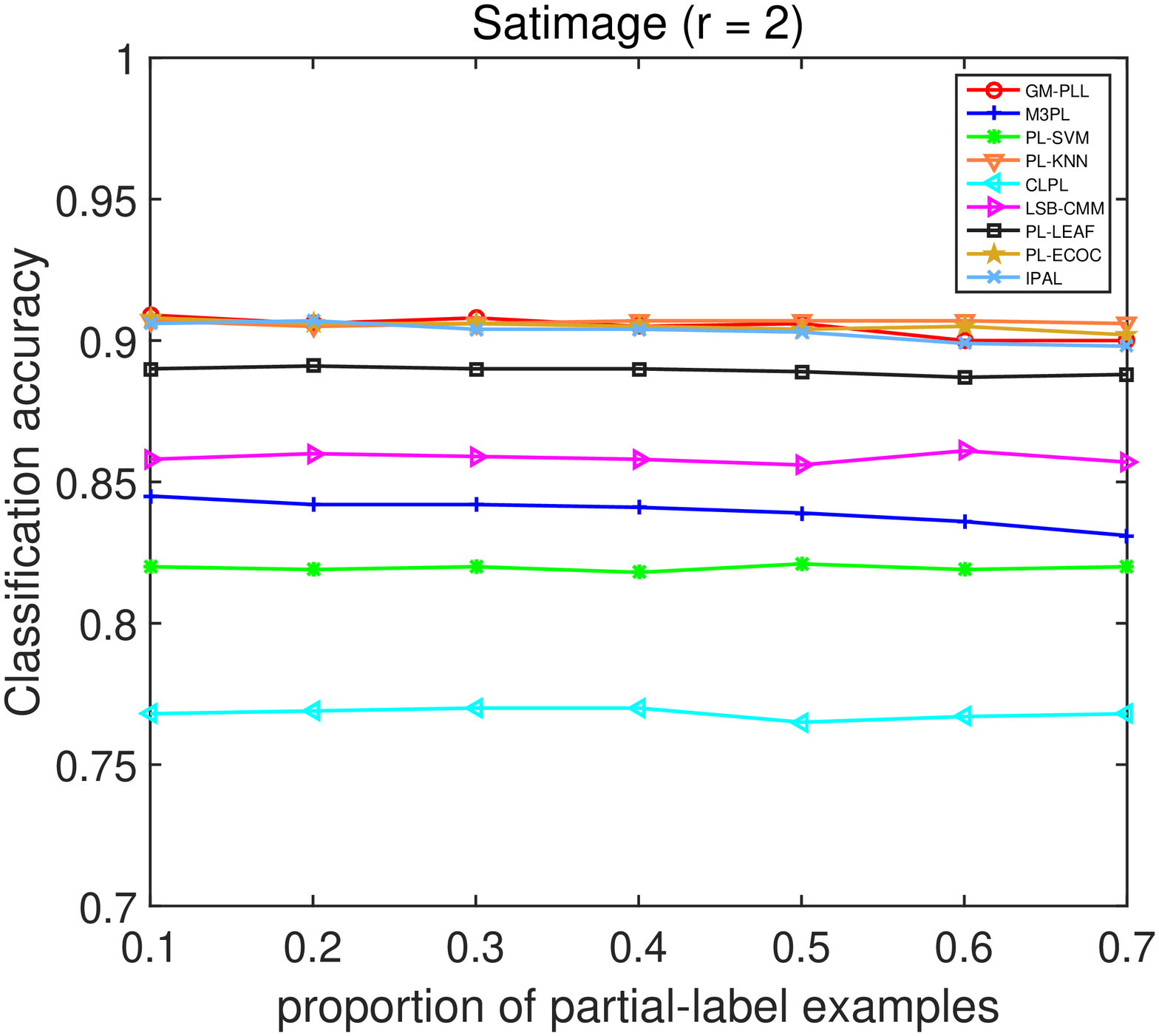}&\includegraphics[width = 2in,height=1.6in]{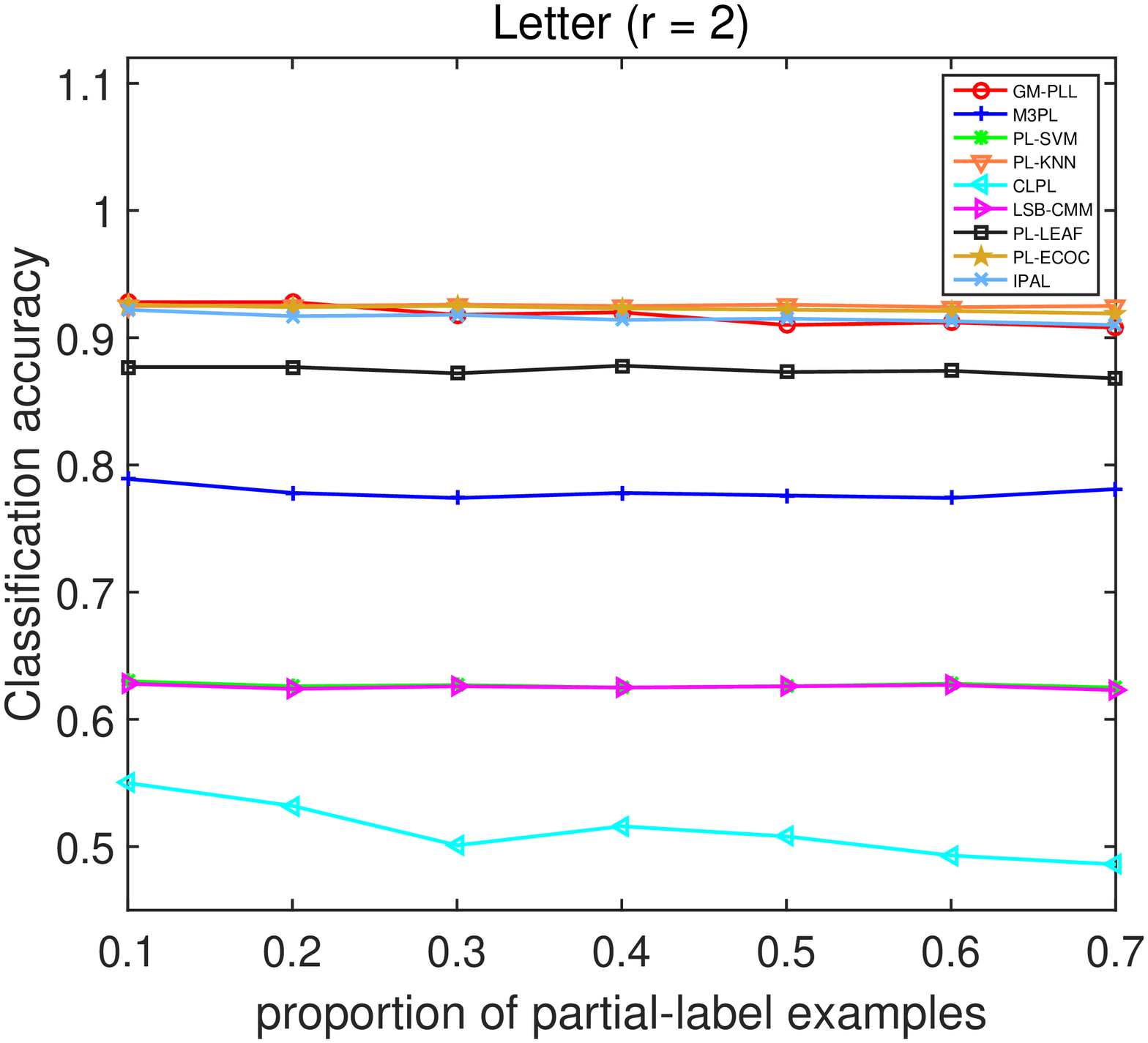}&\includegraphics[width = 2in,height=1.6in]{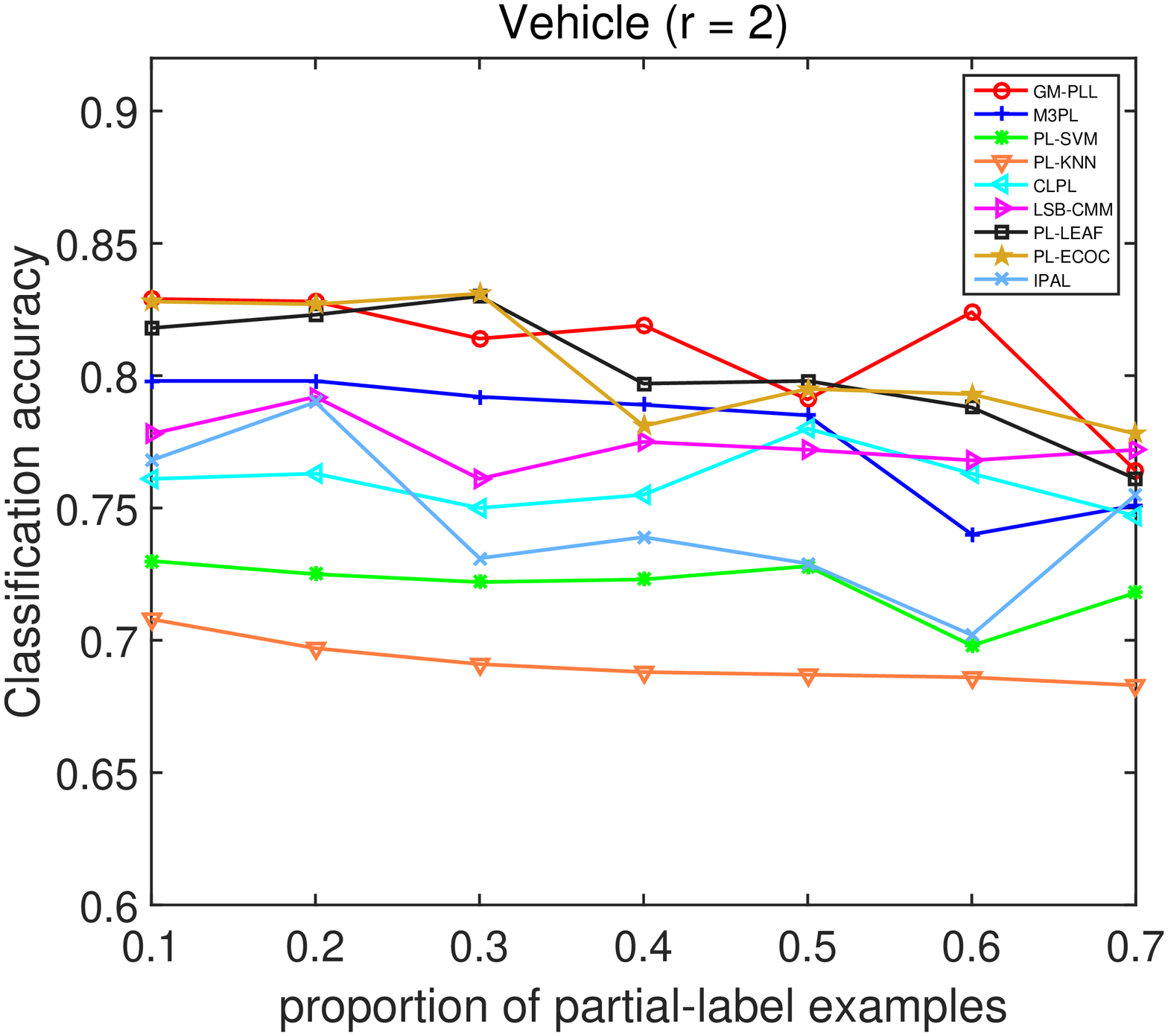}\\
\includegraphics[width = 2in,height=1.6in]{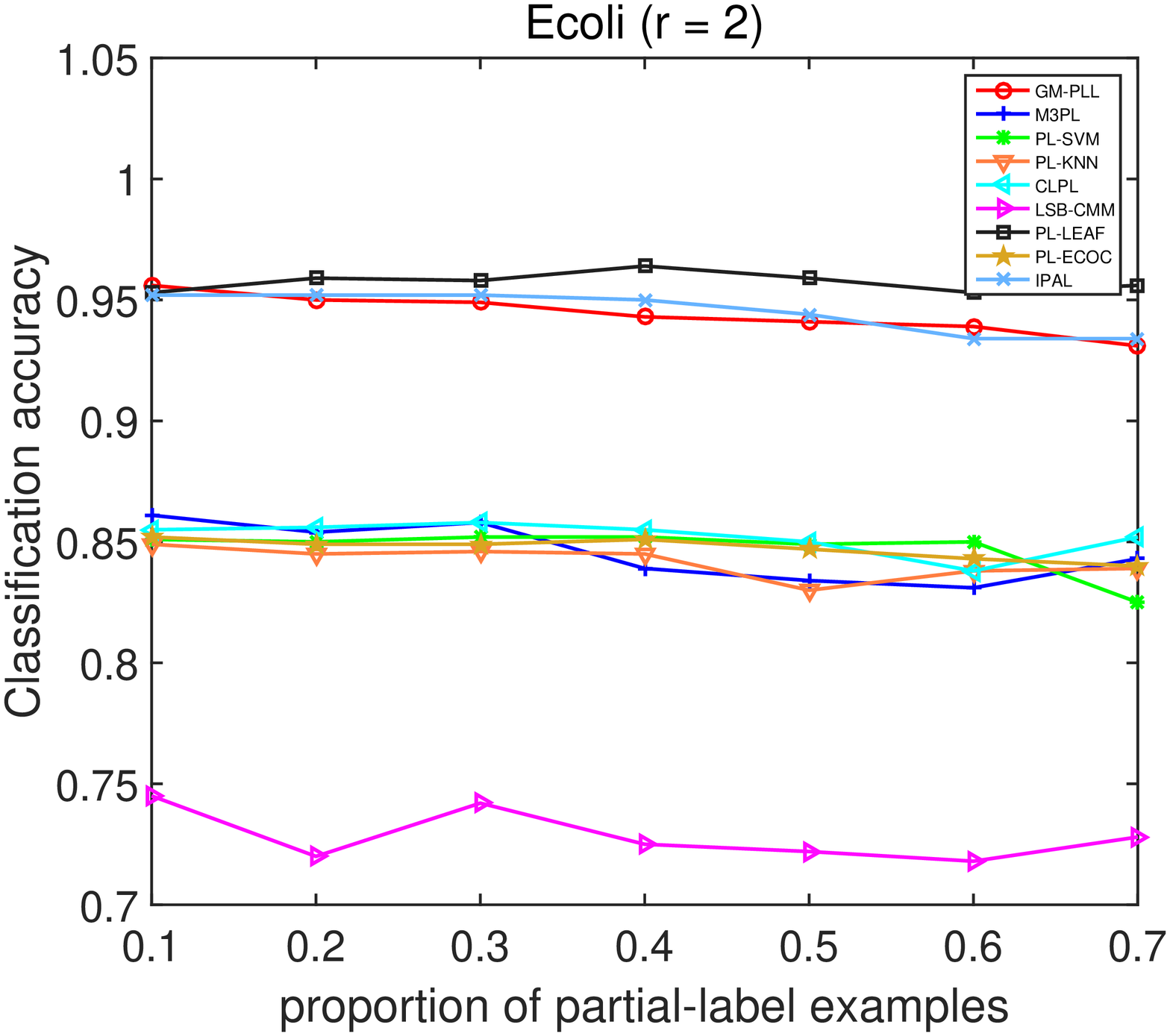}&\includegraphics[width = 2in,height=1.6in]{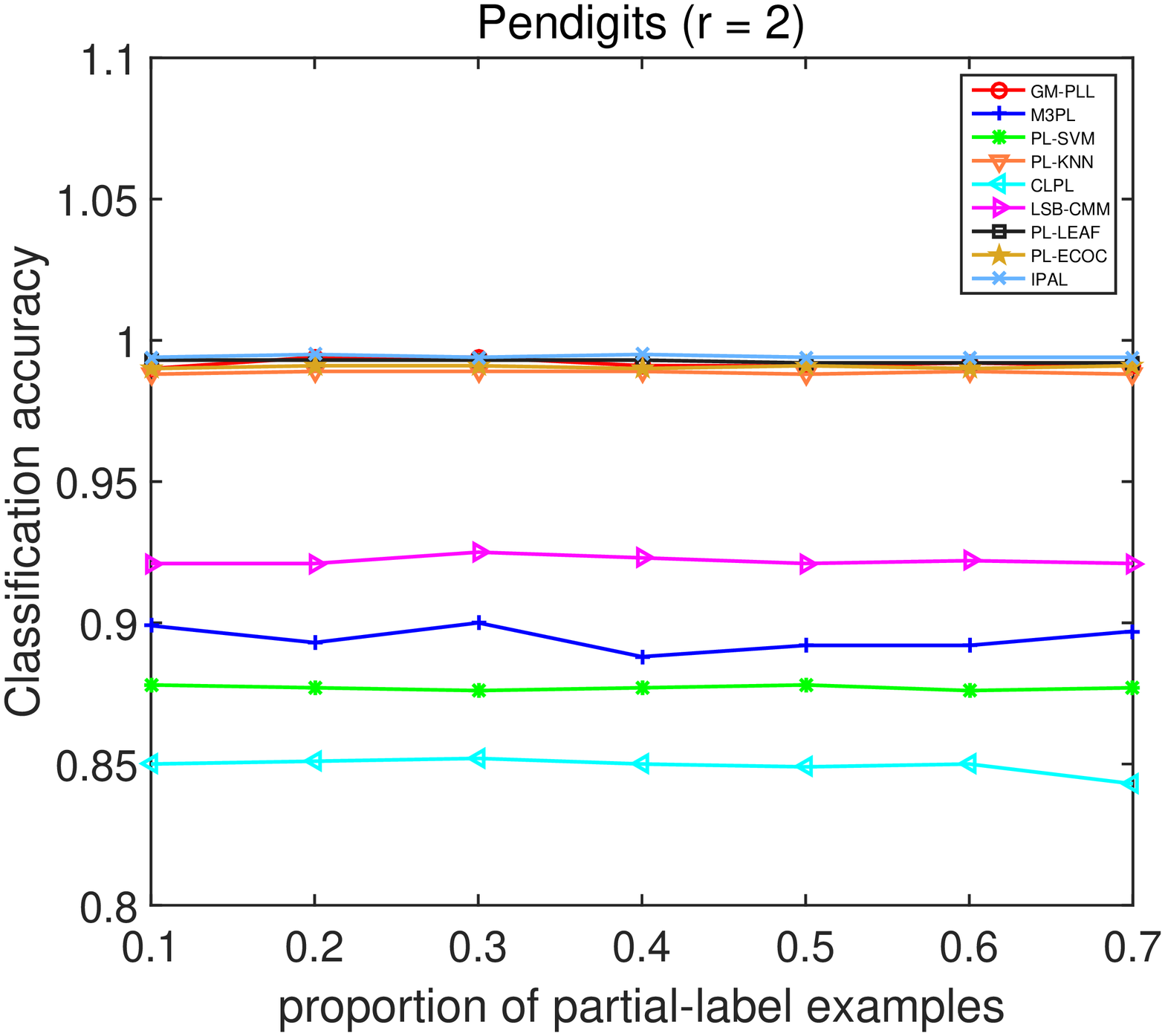}&\includegraphics[width = 2in,height=1.6in]{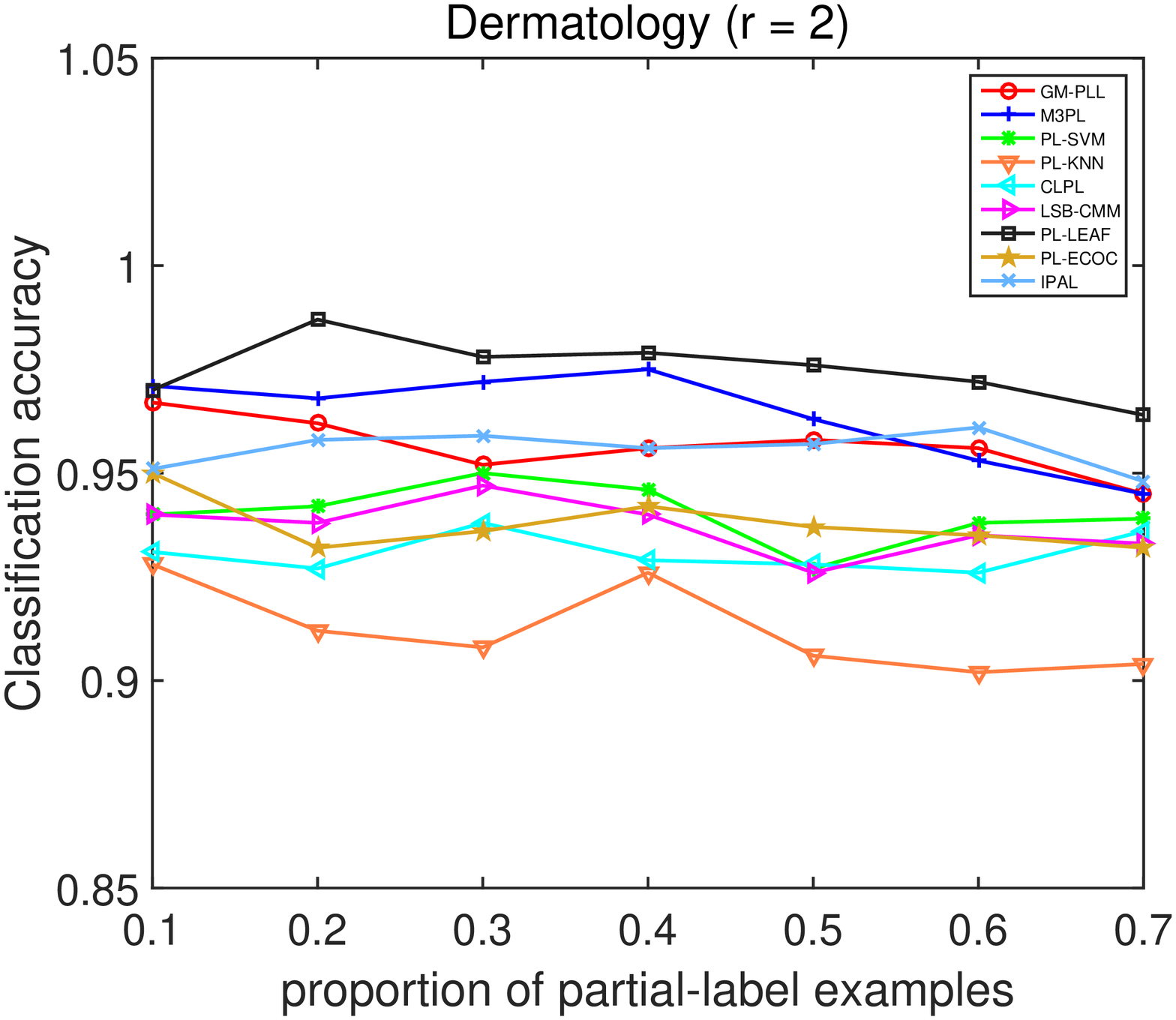}\\
\end{tabular}
\vspace{3mm}
\caption{The classification accuracy of each comparing method on nine controlled UCI data sets with two false positive candidate labels (r = 2)}
\label{Fig-r2}
\vspace{0mm}
\end{figure*}

Meanwhile, we employ \textbf{four} classical (PL-SVM, PL-KNN, CLPL, LSB-CMM) and \textbf{four} state-of-the-art (M3PL, PL-LEAF, PL-ECOC, IPAL) partial label learning algorithms that are based on different disambiguation strategies \footnote{We partially use the open source codes from Zhang Minling's homepage: http://cse.seu.edu.cn/PersonalPage/zhangml/} for comparative studies, where the configured parameters of each method are utilized following the suggestions in respective literatures:

\begin{table*}[!ht]
\centering
\caption{Win/tie/loss counts of the GM-PLL's classification performance against each comparing method on UCI data sets (pairwise $t$-test at 0.05 significance level)}
\vspace{2mm}
\label{wintieloss}
\resizebox{17cm}{!}{
\begin{tabular}{c|cccccccc|c}
\cline{1-6}
\hline \hline
Data set    &PL-KNN  &PL-SVM   &LSB-CMM  &CLPL   & M3PL    &PL-LEAF   &PL-ECOC   &IPAL   &sum  \\\hline
glass       &19/2/0  &21/0/0   &21/0/0   &21/0/0 & 21/0/0  &7/4/10    &21/0/0    &19/2/0 &150/8/10 \\
segment     &21/0/0  &21/0/0   &21/0/0   &21/0/0 & 21/0/0  &21/0/0    &16/5/0    &21/0/0 &163/5/0  \\
vehicle     &21/0/0  &21/0/0   &17/3/1   &18/0/3 & 19/2/0  &8/7/6     &7/5/9     &21/0/0 &132/17/19\\
letter      &14/7/0  &21/0/0   &21/0/0   &21/0/0 & 21/0/0  &21/0/0    &5/16/0    &15/6/0 &139/29/0\\
satimage    &19/2/0  &21/0/0   &21/0/0   &21/0/0 & 21/0/0  &20/1/0    &15/6/0    &19/2/0 &157/11/0\\
abalone     &21/0/0  &21/0/0   &0/0/21   &0/10/11& 21/0/0  &0/0/21    &0/0/21    &21/0/0 &84/10/74\\
ecoli       &12/9/0  &21/0/0   &21/0/0   &21/0/0 & 21/0/0  &1/13/7    &21/0/0    &11/10/0&129/32/7\\
dermatology &14/7/0  &21/0/0   &21/0/0   &21/0/0 & 6/14/1  &0/14/7    &21/0/0    &13/8/0 &117/43/8\\
pendigits   &2/19/0  &21/0/0   &21/0/0   &21/0/0 & 21/0/0  &9/12/0    &11/10/0   &1/20/0 &107/61/0\\ \hline
sum         &163/22/4&178/6/5  &142/0/42 &148/14/27 & 153/15/21 &79/44/66&110/37/42 &125/55/9&-\\\hline \hline
\end{tabular}}
\vspace{-1mm}
\end{table*}

{
\begin{itemize}
\item \textbf{PL-SVM} \cite{Nguyen:cwpl-KDDM2008}: Based on IDS, it gets the predicted-label according to incorporating maximum margin scheme. [suggested configuration: $\lambda\!\in\!{\{10^{-3},10^{-2},\ldots,10^{3}\}}$] ;
\item \textbf{PL-KNN} \cite{Huller:lfale-LNCS2005}: Based on ADS, it obtains the predicted-label according to averaging the outputs of the $k$-nearest neighbors. [suggested configuration: $k$=10];
\item \textbf{CLPL} \cite{Cour:lfpl-JMLR2011}: A convex optimization partial-label learning method based on ADS. [suggested configuration: SVM with hinge loss];
\item \textbf{LSB-CMM} \cite{Liu:acmmmfsll-NIPS2012}: Based on IDS, it makes prediction according to calculating the maximum-likelihood value of the model with unseen instances input. [suggested configuration: \emph{q} mixture components];
\item \textbf{M3PL} \cite{Yu:mmpll-ML2015}: Originated from PL-SVM, it is also based on the maximum-margin strategy, and it gets the predicted-label via calculating the maximum values of model outputs. [suggested configuration: $C_{max}\in{\{10^{-2},10^{-1},\ldots,10^{2}\}}$] ;
\item \textbf{PL-LEAF} \cite{Zhang:pllvfad-TKDD2016}: A partial-label learning method via feature-aware disambiguation. [suggested configuration: $k$=10, $C_1=10$, $C_2=1$];
\item \textbf{IPAL} \cite{Zhang:stpllpaiba-IJCAI2015}: it disambiguates the candidate label set by taking the instance similarity into consideration. [suggested configuration: $k$=10];
\item \textbf{PL-ECOC} \cite{zhang:dfpll-IEEET2017}: Based on a coding-decoding procedure, it learns from partial-label training examples in a disambiguation-free manner. [suggested configuration: the codeword length $L = \lceil\emph{log}_{2}(q)\rceil$];
\end{itemize}}

Before conducting the experiments, we give the range of the required variables. In detail, during the training phase, the threshold variable $\beta$ is set among $\{0.3, 0.4, \ldots, 0.8\}$ to exploit the most valuable similarity information and dissimilarity information. And the coefficient parameter $\alpha$ is chosen from $\{0, 0.1, 0.2\}$ to balance the effect of the number of varying label categories. During the test phase, inspired by \cite{Zhang:stpllpaiba-IJCAI2015}, we empirically set ${k} = 10$ for $k$-nearest neighbor instances to complete the candidate label set of each unseen instance, and meanwhile the size of the label set $r$ is empirically set to more than 1 to guarantee that the ground-truth label can be involved in the assigned candidate label set. After initializing the above variables, we adopt ten-fold cross-validation to train the model and get the average classification accuracy on each data set.

\begin{figure*}[!ht]
\centering
\begin{tabular}{ccc}
\includegraphics[width = 2in,height=1.6in]{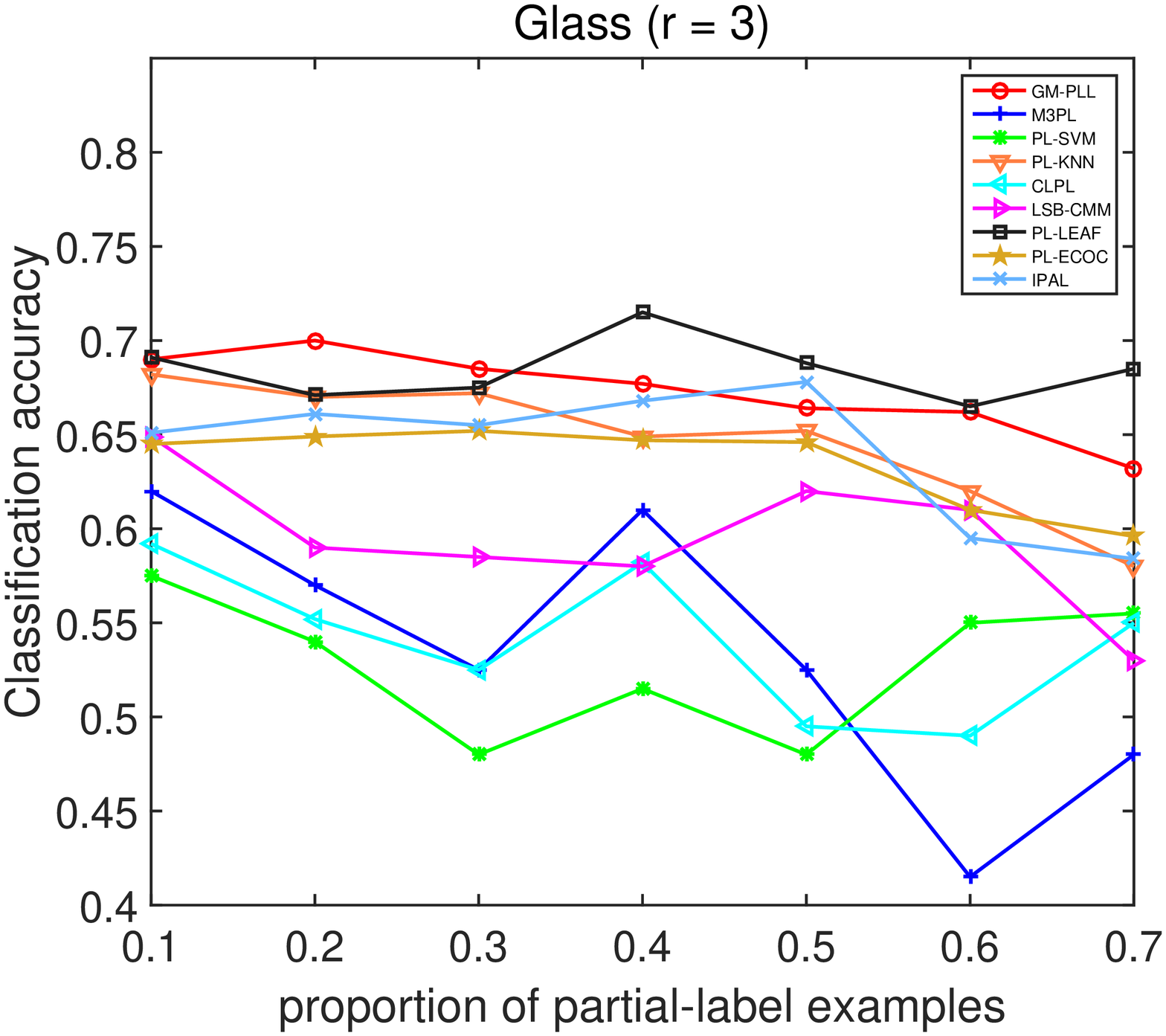}&\includegraphics[width = 2in,height=1.6in]{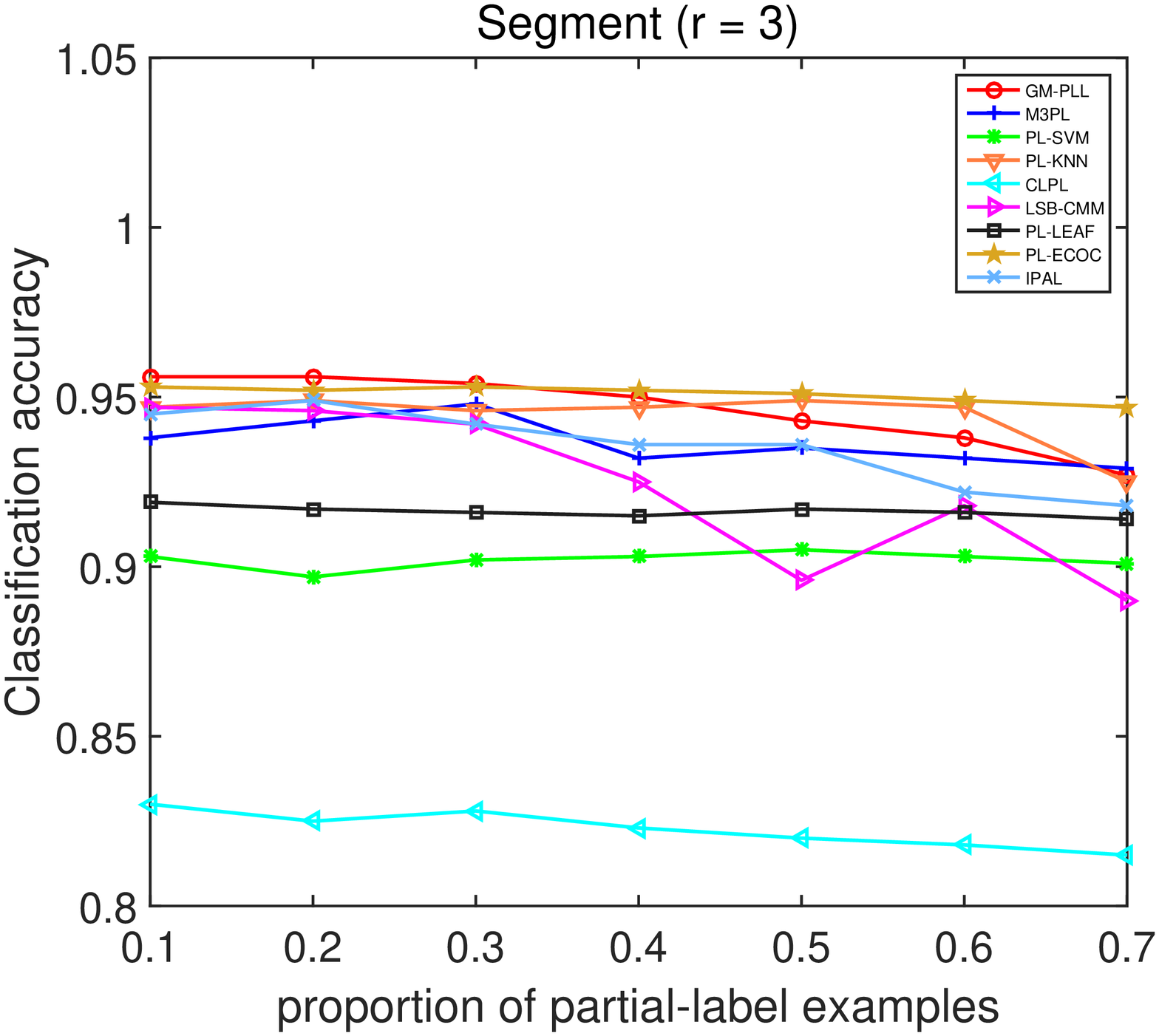}&\includegraphics[width = 2in,height=1.6in]{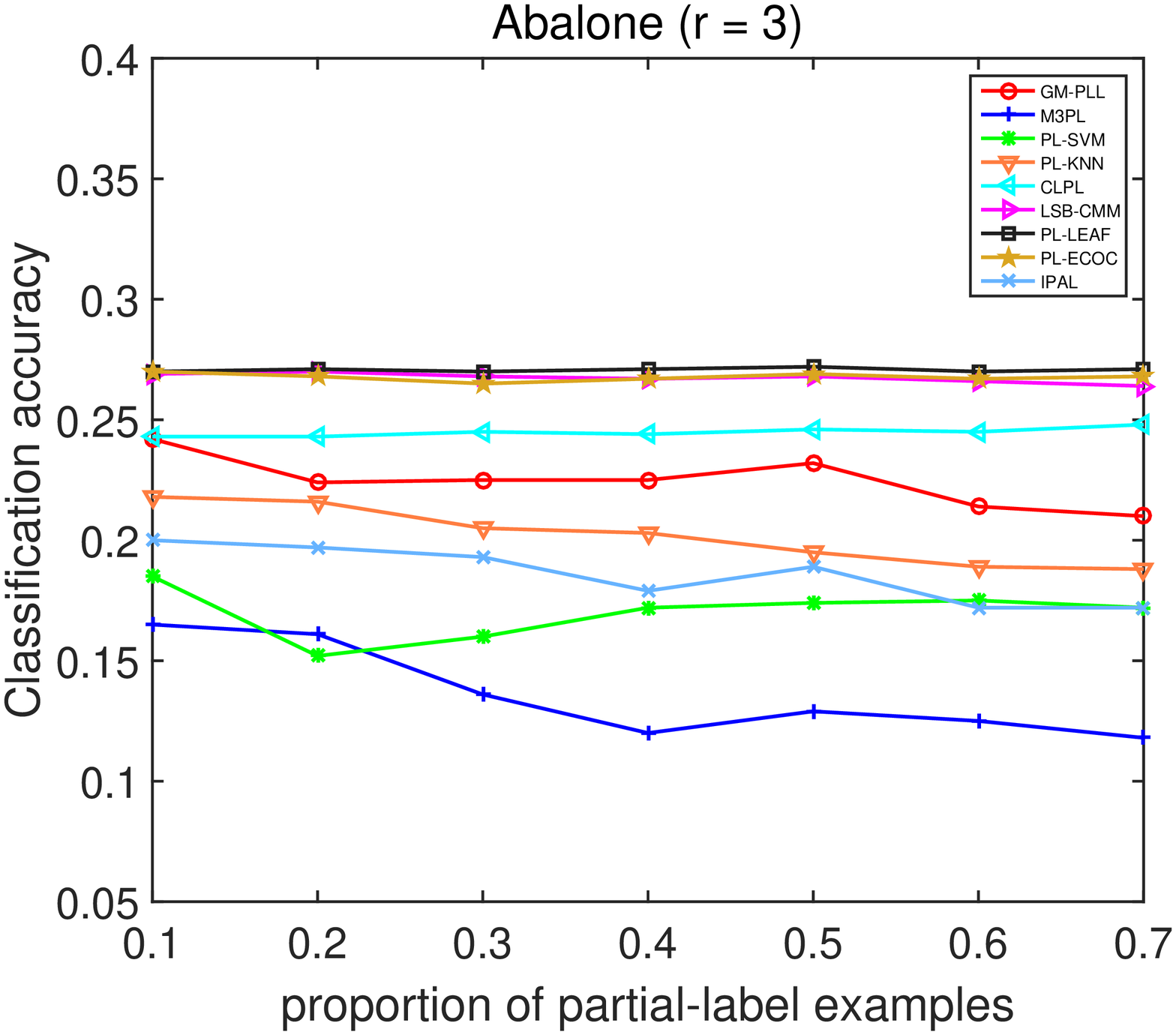}\\
\includegraphics[width = 2in,height=1.6in]{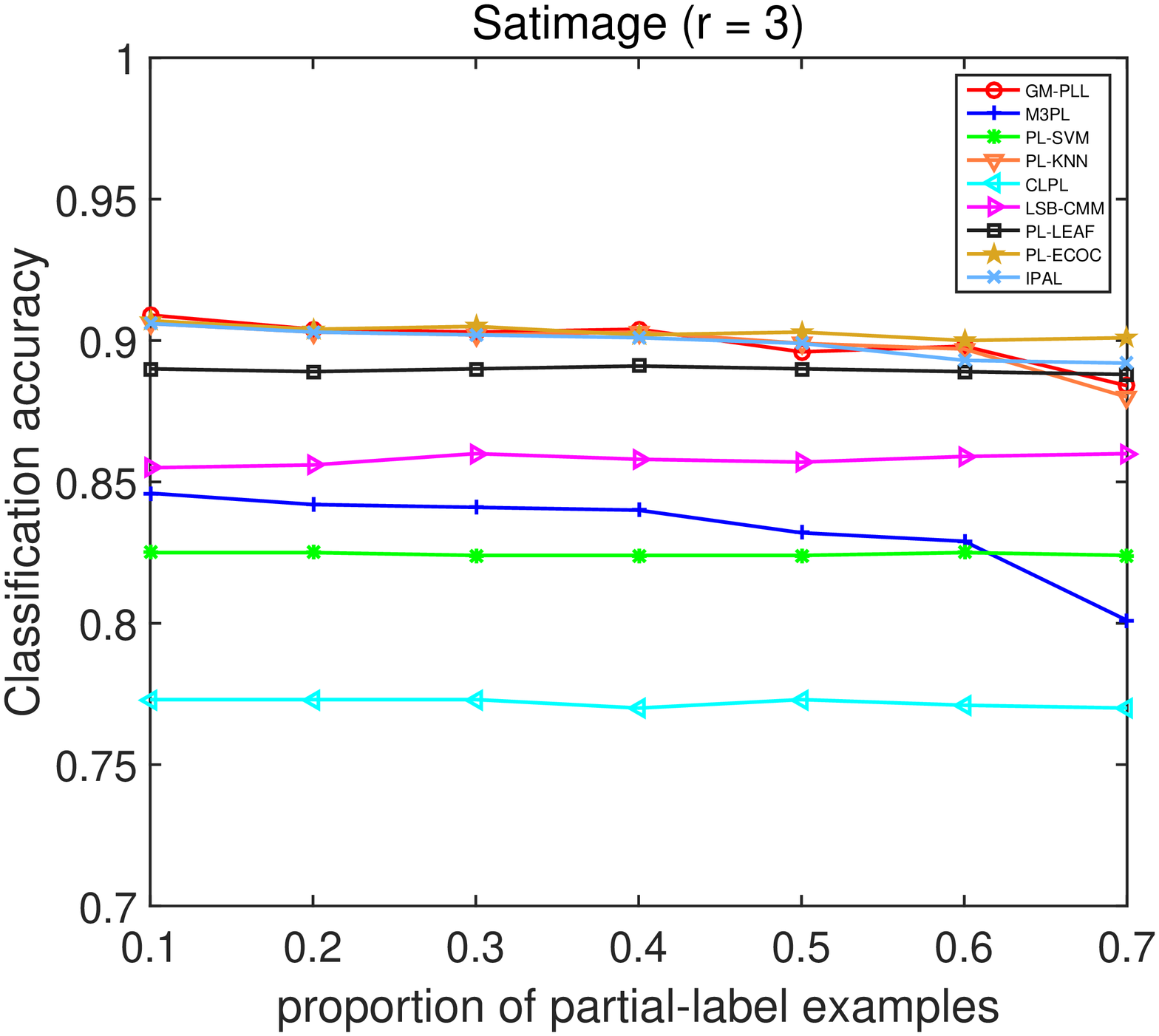}&\includegraphics[width = 2in,height=1.6in]{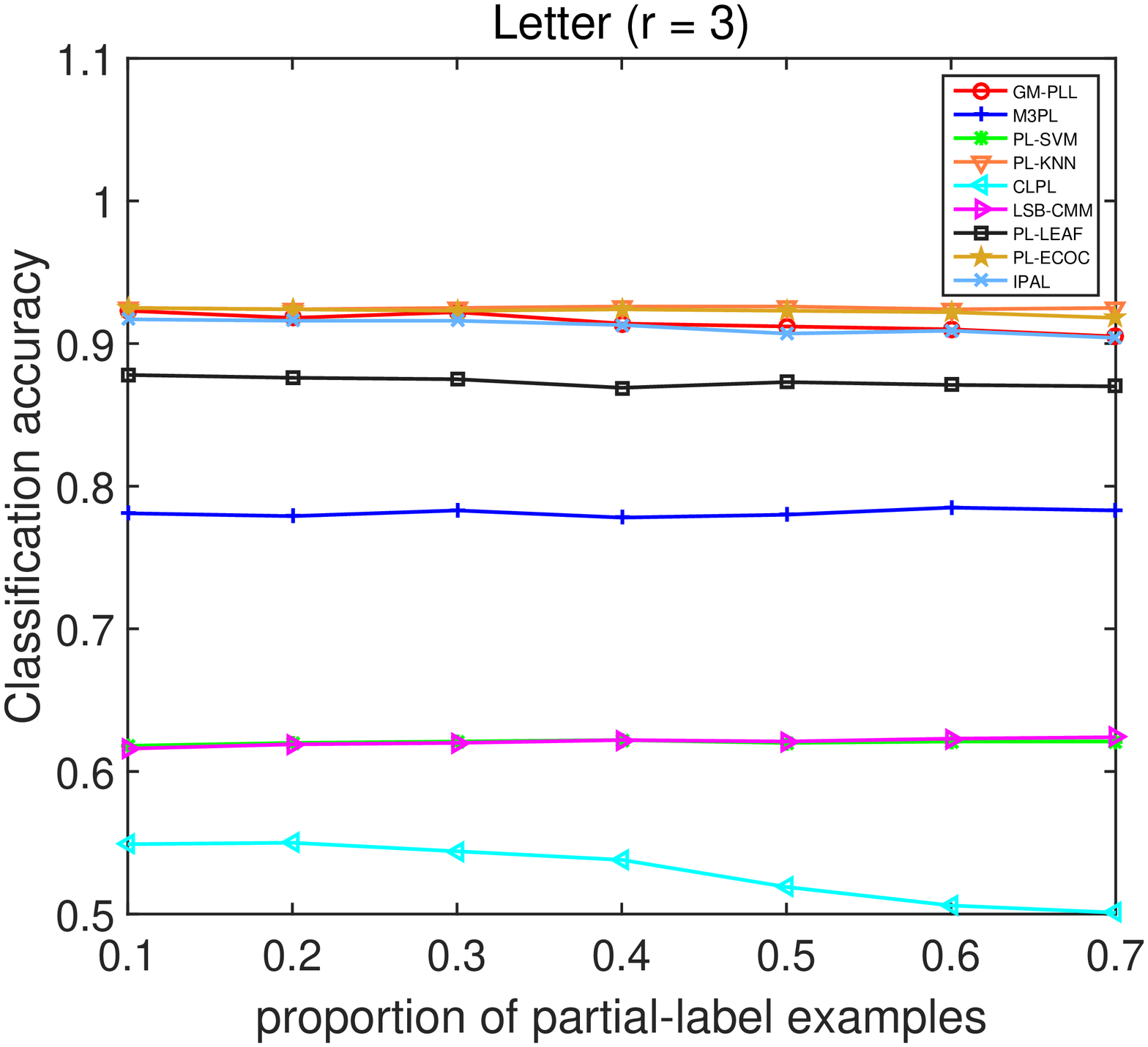}&\includegraphics[width = 2in,height=1.6in]{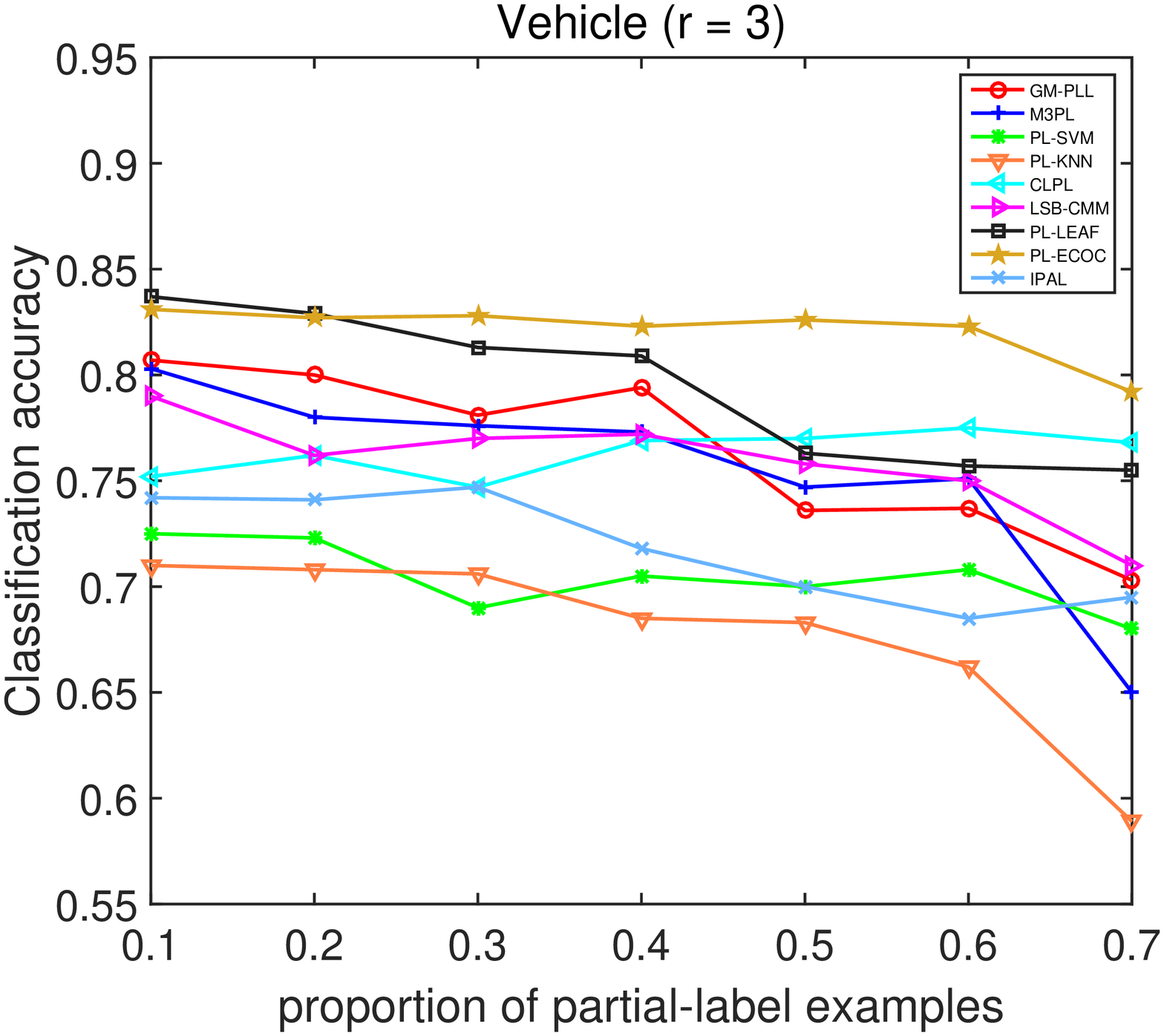}\\
\includegraphics[width = 2in,height=1.6in]{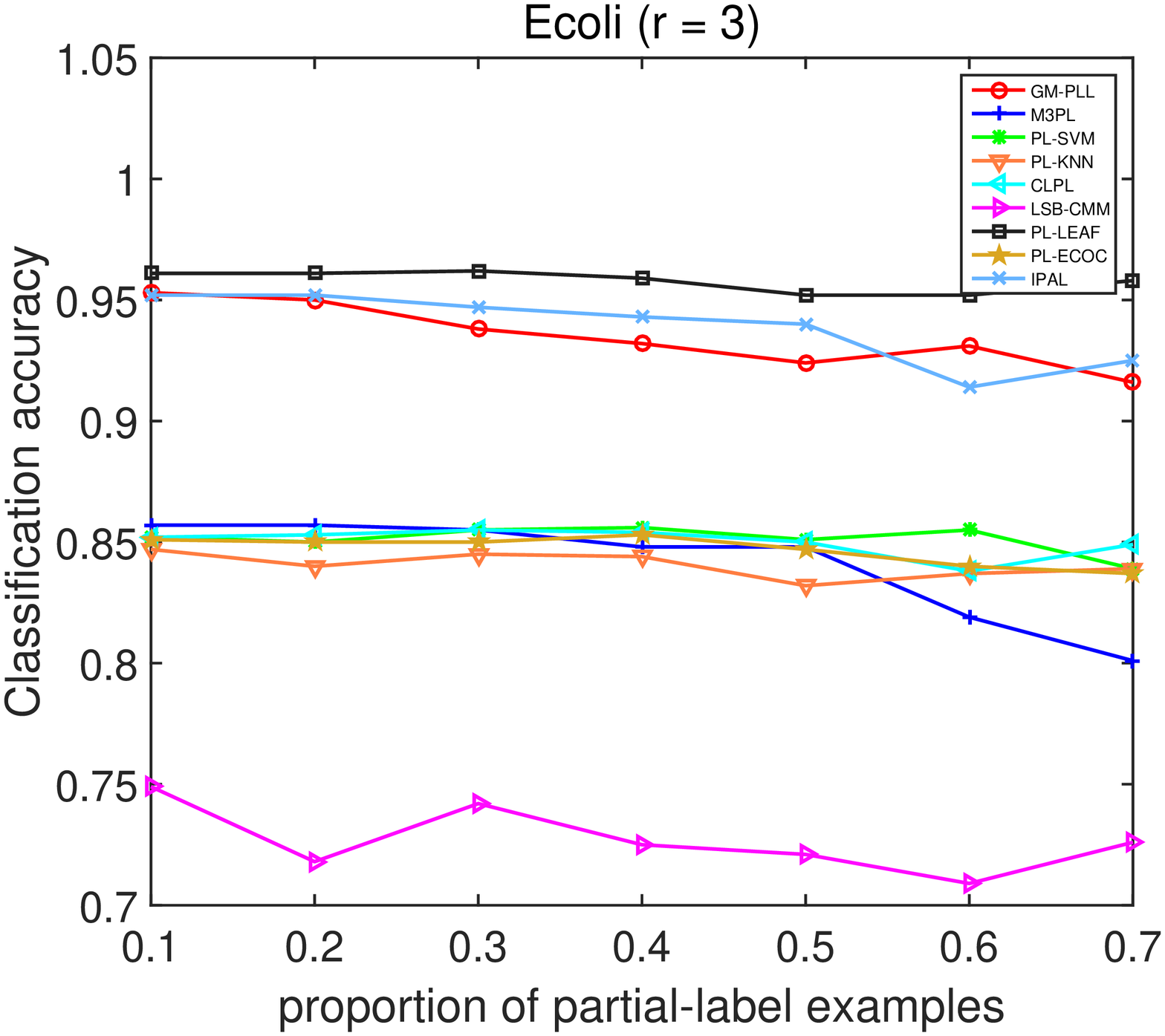}&\includegraphics[width = 2in,height=1.6in]{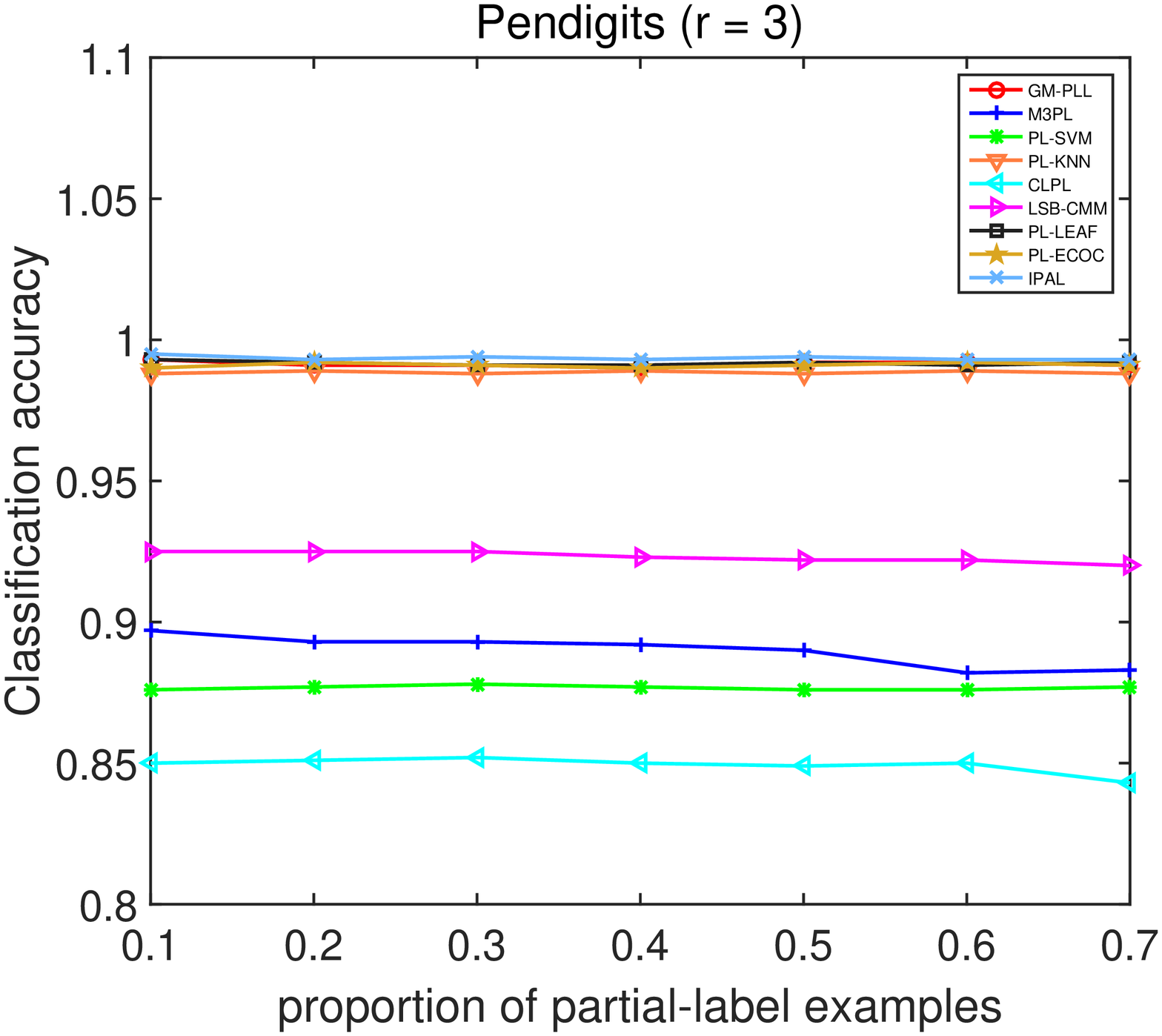}&\includegraphics[width = 2in,height=1.6in]{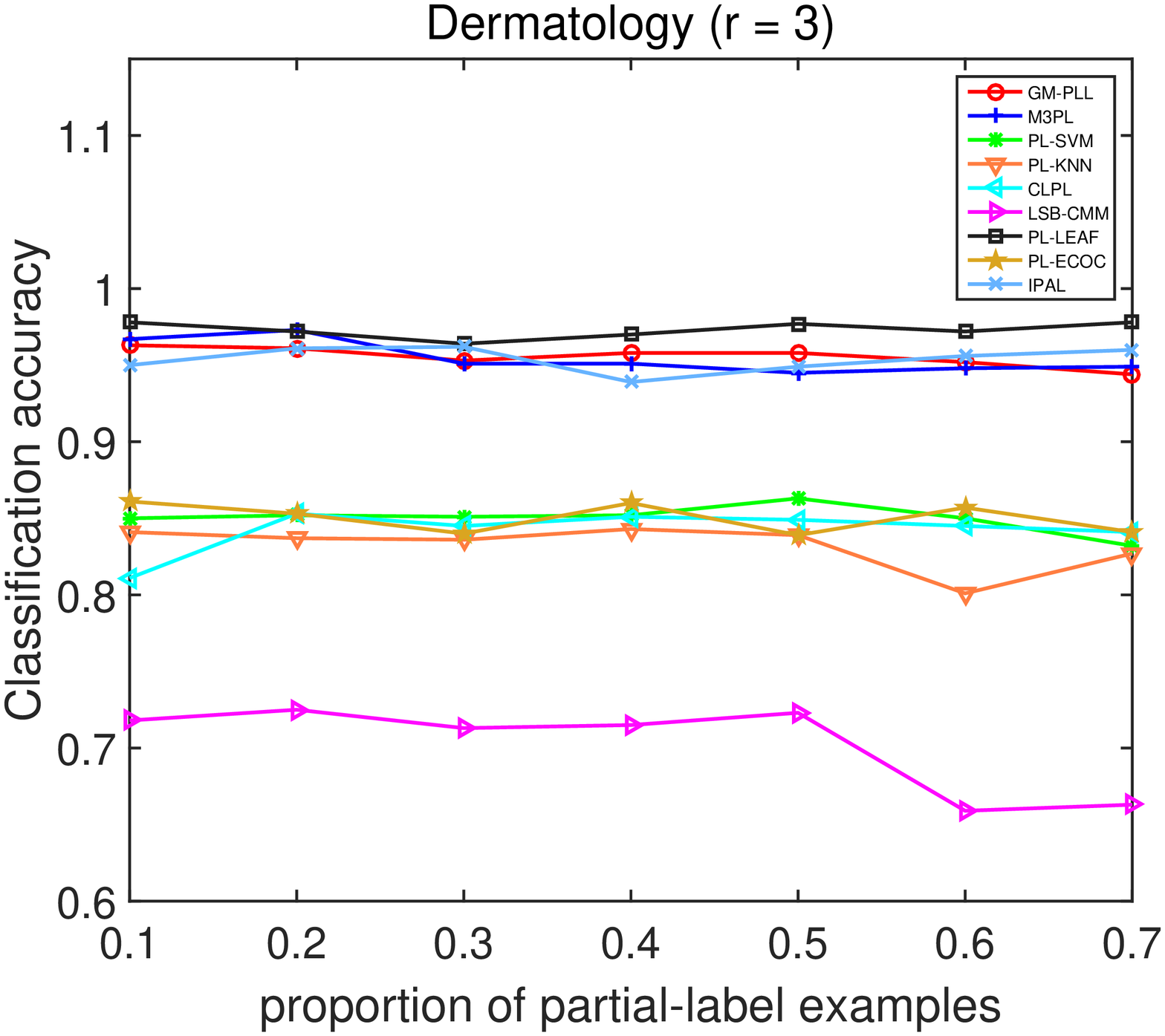}\\
\end{tabular}
\vspace{3mm}
\caption{The classification accuracy of each comparing method on nine controlled UCI data sets with three false positive candidate labels (r = 3)}
\label{Fig-r3}
\vspace{0mm}
\end{figure*}

\begin{table*}[!ht]
\centering
\setlength{\abovecaptionskip}{0pt}%
\setlength{\belowcaptionskip}{5pt}%
\caption{ Inductive accuracy (mean $\pm$ std) of each comparing algorithm on real-world data sets. $\bullet/\circ$ indicates that GM-PLL is statistically superior / inferior to the comparing algorithm on each data set (pairwise $t$-text at 0.05 significate level). }
\label{Table3}
\vspace{2mm}
\resizebox{17cm}{!}{
\begin{tabular}{ccccccc}
\toprule[0.5pt]
\toprule[0.5pt]
 & Lost  & MSRCv2 & Yahoo! News & BirdSong & SoccerPlayer & FG-NET
 \vspace{1mm} \\
\toprule[0.5pt]
GM-PLL  & \textbf{0.737$\pm$0.043}   & \textbf{0.530$\pm$0.019}    & 0.629$\pm$0.007    & 0.663$\pm$0.010& \textbf{0.549$\pm$0.009}          & 0.065$\pm$0.021   \\
\toprule[1.0pt]
PL-SVM  & 0.639$\pm$0.056 $\bullet$ & 0.417$\pm$0.027 $\bullet$  & 0.636$\pm$0.018 $\circ$  & 0.662$\pm$0.032 $\bullet$   & 0.430$\pm$0.004 $\bullet$        & 0.058$\pm$0.010 $\bullet$ \\
CLPL    & 0.670$\pm$0.024 $\bullet$ & 0.375$\pm$0.020 $\bullet$  & 0.462$\pm$0.009 $\bullet$   & 0.632$\pm$0.017 $\bullet$   & 0.347$\pm$0.004 $\bullet$        & 0.047$\pm$0.017 $\bullet$  \\
PL-KNN  & 0.332$\pm$0.030 $\bullet$ & 0.417$\pm$0.012 $\bullet$  & 0.457$\pm$0.009 $\bullet$    & 0.614$\pm$0.024 $\bullet$   & 0.494$\pm$0.004 $\bullet$       & 0.037$\pm$0.008 $\bullet$  \\
LSB-CMM & 0.591$\pm$0.019 $\bullet$ & 0.431$\pm$0.008 $\bullet$  & 0.648$\pm$0.015 $\circ$    & 0.717$\pm$0.024 $\circ$   & 0.506$\pm$0.006 $\bullet$        & 0.056$\pm$0.008 $\bullet$  \\
\toprule[1.0pt]
M3PL    & 0.732$\pm$0.035 $\bullet$ & 0.521$\pm$0.030 $\bullet$  & 0.655$\pm$0.010 $\circ$    & 0.709$\pm$0.010 $\circ$   & 0.446$\pm$0.013 $\bullet$        & 0.037$\pm$0.025 $\bullet$  \\
PL-LEAF & 0.664$\pm$0.020 $\bullet$ & 0.459$\pm$0.013 $\bullet$  & 0.597$\pm$0.012 $\bullet$    & 0.706$\pm$0.012 $\circ$   & 0.515$\pm$0.004 $\bullet$        & \textbf{0.072$\pm$0.010} $\circ$  \\
IPAL    & 0.726$\pm$0.041 $\bullet$ & 0.523$\pm$0.025 $\bullet$  & \textbf{0.667$\pm$0.014} $\circ$    & 0.708$\pm$0.014 $\circ$   & 0.547$\pm$0.014 $\bullet$        & 0.057$\pm$0.023 $\bullet$  \\
PL-ECOC & 0.703$\pm$0.052 $\bullet$ & 0.505$\pm$0.027 $\bullet$  & 0.662$\pm$0.010 $\circ$    & 0.740$\pm$0.016 $\circ$   & 0.537$\pm$0.020 $\bullet$        &   0.040$\pm$0.018  $\bullet$     \\
\toprule[0.5pt]
\toprule[0.5pt]
\end{tabular}}
\vspace{-2mm}
\end{table*}

\subsection{Experimental Results}

Since the origins of the two kinds of data sets are different, nine UCI data sets are constructed manually while six RW data sets come from real world scenarios, we conduct two series of experiments to evaluate the proposed method and the experimental results are exhibited in the following two subsections separately. In our paper, the experimental results of the comparing algorithms originate from two aspects: one is from the results we implemented by utilizing the source codes provided by the authors; the other is from the results exhibited in the respective literatures.

\begin{table*}[!ht]
\centering
\setlength{\abovecaptionskip}{0pt}%
\setlength{\belowcaptionskip}{5pt}%
\caption{ Transductive accuracy (mean $\pm$ std) of each comparing algorithm on real-world data sets. $\bullet/\circ$ indicates that GM-PLL is statistically superior / inferior to the comparing algorithm on each data set (pairwise $t$-text at 0.05 significate level). }
\label{Tabledisambiguation}
\vspace{2mm}
\resizebox{17cm}{!}{
\begin{tabular}{ccccccc}
\toprule[0.5pt]
\toprule[0.5pt]
 & Lost  & MSRCv2 & Yahoo! News & BirdSong & SoccerPlayer & FG-NET
 \vspace{1mm} \\
\toprule[0.5pt]
GM-PLL  & 0.881$\pm$0.005   & \textbf{0.770$\pm$0.013}    & 0.705$\pm$0.612    & 0.834$\pm$0.010 & 0.668$\pm$0.003         & \textbf{0.186$\pm$0.021}   \\
\toprule[1.0pt]
PL-SVM  & 0.887$\pm$0.012 $\circ$ & 0.653$\pm$0.024 $\bullet$  & 0.871$\pm$0.002 $\circ$    & 0.825$\pm$0.012$\bullet$    & 0.688$\pm$0.014 $\circ$        & 0.136$\pm$0.021 $\bullet$ \\
CLPL    & \textbf{0.894$\pm$0.005} $\circ$ & 0.656$\pm$0.010 $\bullet$  & 0.834$\pm$0.002 $\circ$    & 0.822$\pm$0.004$\bullet$    & 0.680$\pm$0.010 $\bullet$        & 0.158$\pm$0.018 $\bullet$  \\
PL-KNN  & 0.615$\pm$0.036 $\bullet$ & 0.616$\pm$0.006 $\bullet$  & 0.692$\pm$0.010 $\bullet$    & 0.772$\pm$0.021$\bullet$    & 0.492$\pm$0.015 $\bullet$       & 0.173$\pm$0.017 $\bullet$  \\
LSB-CMM & 0.721$\pm$0.010 $\bullet$ & 0.524$\pm$0.007 $\bullet$  & \textbf{0.872$\pm$0.001} $\circ$    & 0.716$\pm$0.014$\bullet$    & 0.704$\pm$0.002 $\circ$        & 0.138$\pm$0.019 $\bullet$  \\
\toprule[1.0pt]
M3PL    & 0.860$\pm$0.006 $\bullet$ & 0.732$\pm$0.025 $\bullet$  & 0.870$\pm$0.002 $\circ$    & 0.855$\pm$0.030$\circ$    & \textbf{0.761$\pm$0.010} $\circ$        & 0.127$\pm$0.013 $\bullet$  \\
PL-LEAF & 0.809$\pm$0.022 $\bullet$ & 0.645$\pm$0.015 $\bullet$  & 0.827$\pm$0.002 $\circ$    & 0.882$\pm$0.014$\circ$    & 0.702$\pm$0.003 $\circ$        & 0.148$\pm$0.009 $\bullet$  \\
IPAL    & 0.840$\pm$0.041 $\bullet$ & 0.714$\pm$0.015 $\bullet$  & 0.823$\pm$0.008 $\circ$    & 0.833$\pm$0.030$\bullet$    & 0.673$\pm$0.014 $\bullet$        & 0.158$\pm$0.024 $\bullet$  \\
PL-ECOC & 0.851$\pm$0.013 $\bullet$ & 0.555$\pm$0.030 $\bullet$  & 0.862$\pm$0.007 $\circ$    & \textbf{0.886$\pm$0.014}$\circ$    & 0.671$\pm$0.003 $\bullet$        &   0.132$\pm$0.019  $\bullet$     \\
\toprule[0.5pt]
\toprule[0.5pt]
\end{tabular}}
\vspace{-2mm}
\end{table*}

\subsubsection{\textbf{Controlled UCI data sets}}
Figure \ref{Fig-r1}-\ref{Fig-r3} illustrate the classification accuracy of each comparing method on the nine controlled data sets as $p$ increases from $0.1$ to $0.7$ with the step-size $0.1$. Together with the ground-truth label, the $r$ class labels are randomly chosen from $\mathcal{Y}$ to constitute the rest of each candidate label set, where $r = 1,2,3$. Table \ref{wintieloss} summaries the win/tie/loss counts between GM-PLL and other comparing methods. Out of 189 (9 data sets $\times$ 21 configurations) statistical comparisons show that GM-PLL achieves either superior or comparable performance against the eight comparing methods, which is embodied in the following aspects:

\begin{itemize}
\item Among the comparing methods, GM-PLL achieves superior performance against PL-KNN, PL-SVM, LSB-CMM, CLPL and M3PL in most cases. And compared with PL-LEAF, PL-ECOC and IPAL, it also achieves superior or comparable performance in 65.08\%, 77.78\%, 95.23\% cases, respectively. These results demonstrate that the proposed method has superior capacity of disambiguation against other methods based on varying disambiguation strategies, as well as disambiguation-free strategy.
\item Compared with the methods that directly establish INS-GTL assignments, GM-PLL achieves superior performance on most data sets. For example, the average classification accuracy of GM-PLL is 11.2\% higher than M3PL on \emph{Glass} data set and 29.5\% higher than PL-SVM on \emph{Satimage} data set. Meanwhile, GM-PLL also has higher or comparable classification accuracy against the comparing state-of-the-art methods on other controlled UCI data sets. We attribute such success to that it can utilize the co-occurrence possibility of varying instance-label assignments to obtain the accurate INS-GTL assignments.
\item Compared with the methods utilizing the instance similarity, GM-PLL also achieves competitive performance. From the perspective of the Average Classification Accuracy, GM-PLL gets 1.2\% higher than IPAL on \emph{Segment} data set and 1.4\% higher than PL-LEAF on \emph{Letter} data set, respectively; And from the perspective of the Max-Min of classification accuracy, GM-PLL is only 0.84\% higher on \emph{Glass} data set while all other methods are more than 1\%. Moreover, the standard deviation of GM-PLL classification accuracy is lower than the other comparing methods on most data sets. These results clearly indicate the advantage of the proposed method against other instance-similarity based methods.
\end{itemize}

\begin{figure*}[!ht]
\centering
\begin{tabular}{ccc}
\includegraphics[width = 2in,height=1.6in]{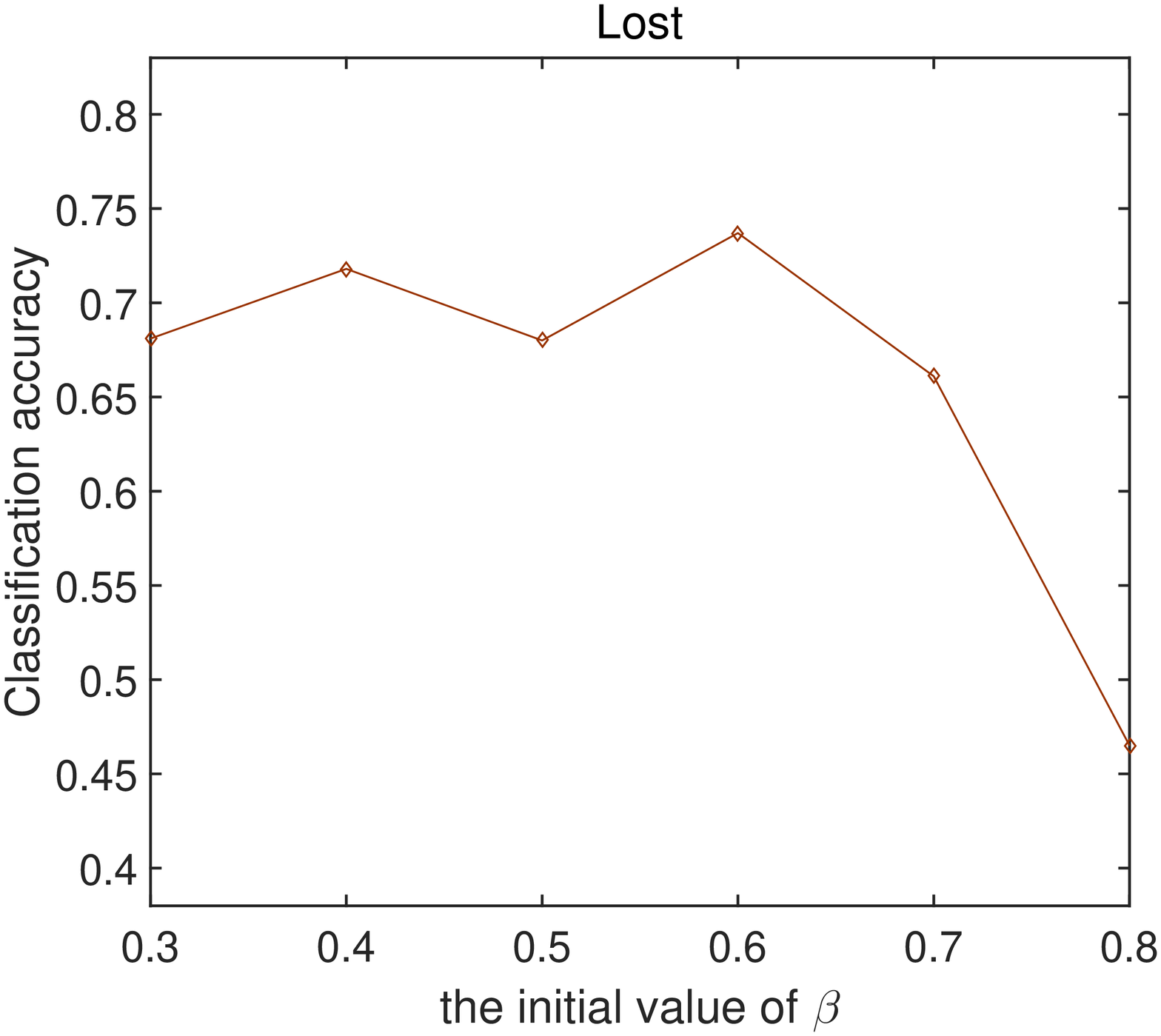}&\includegraphics[width = 2in,height=1.6in]{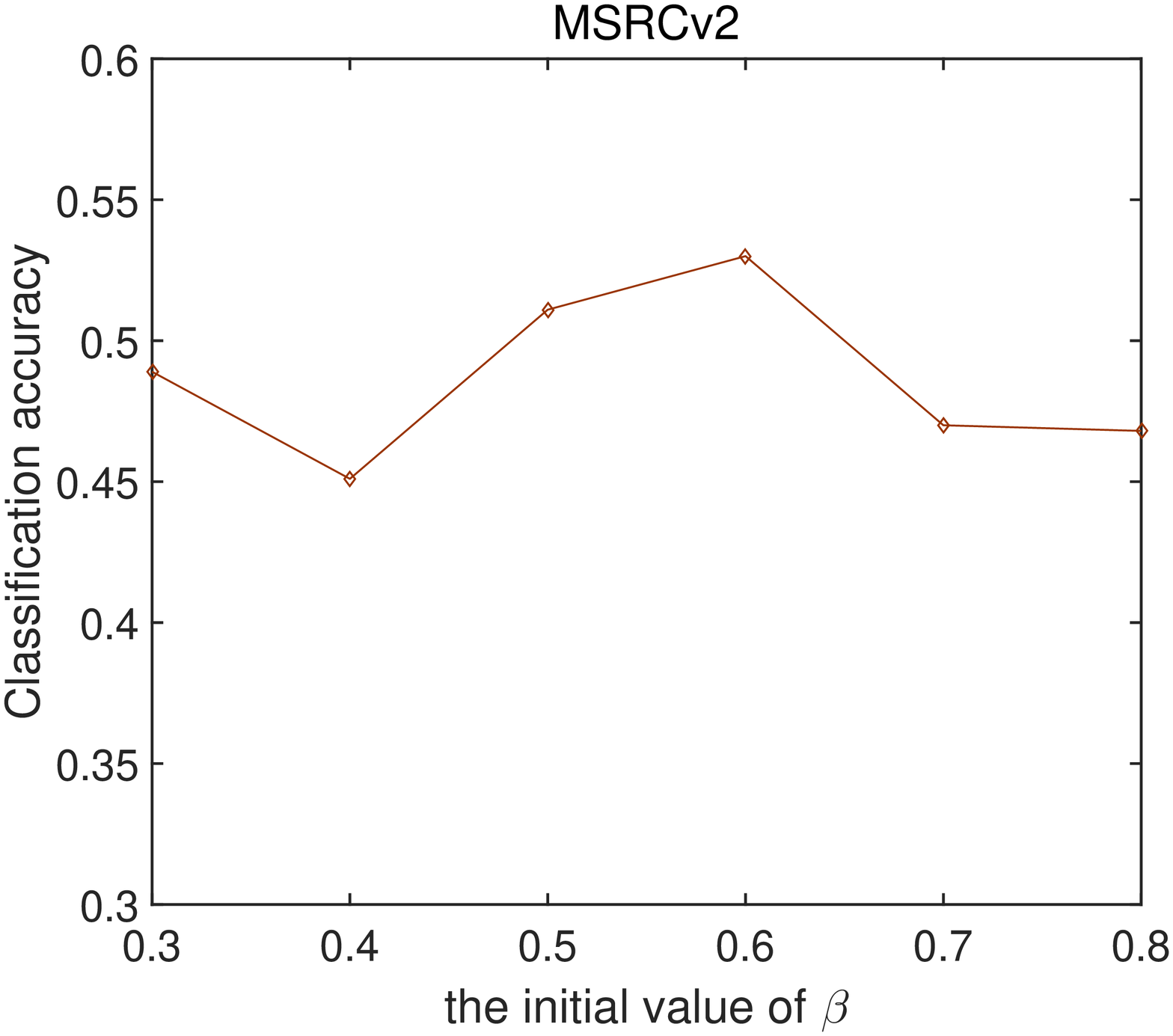}&\includegraphics[width = 2in,height=1.6in]{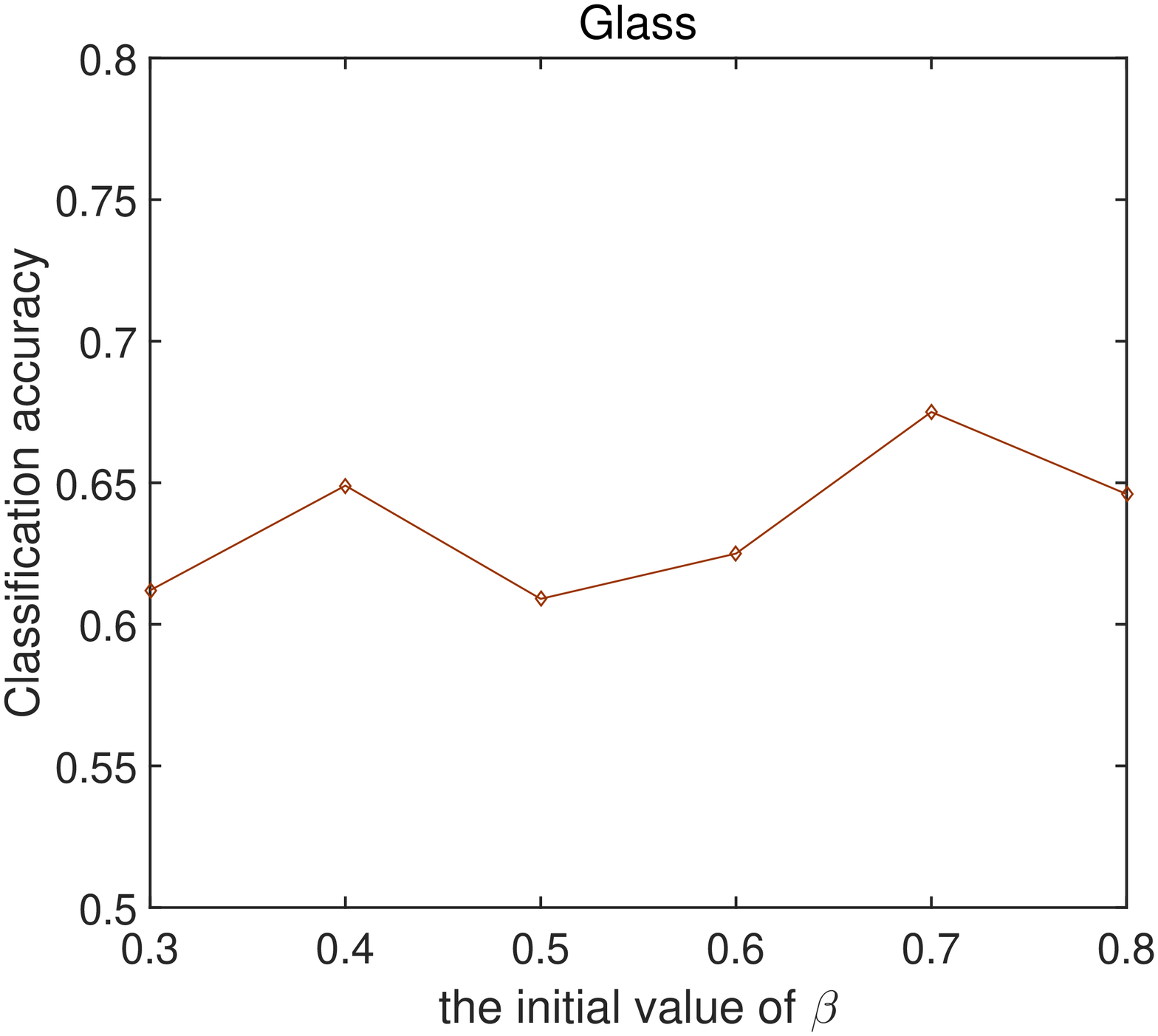}\\
\end{tabular}
\vspace{3mm}
\caption{The classification accuracy of the proposed methods on \emph{Lost}, \emph{MSRCv2} and \emph{Glass} data sets with $r$ fixed ($r = 3$ on \emph{Lost} data set, $r = 4$ on \emph{MSRCv2} data set and $r = 4$ on \emph{Glass} data set respectively)}
\label{Fig-analy1}
\vspace{0mm}
\end{figure*}

\begin{figure*}[!ht]
\centering
\begin{tabular}{ccc}
\includegraphics[width = 2in,height=1.6in]{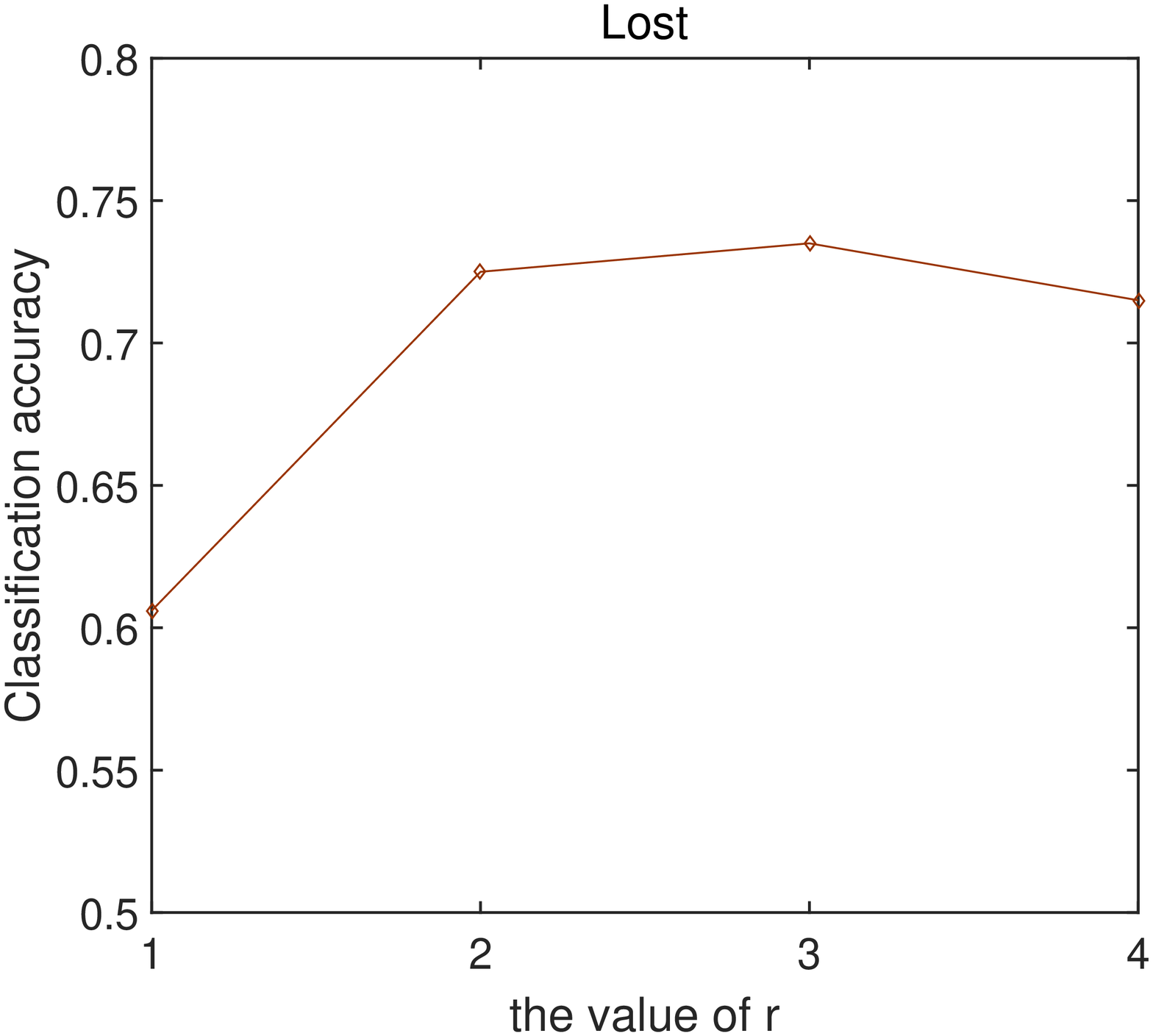}&\includegraphics[width = 2in,height=1.6in]{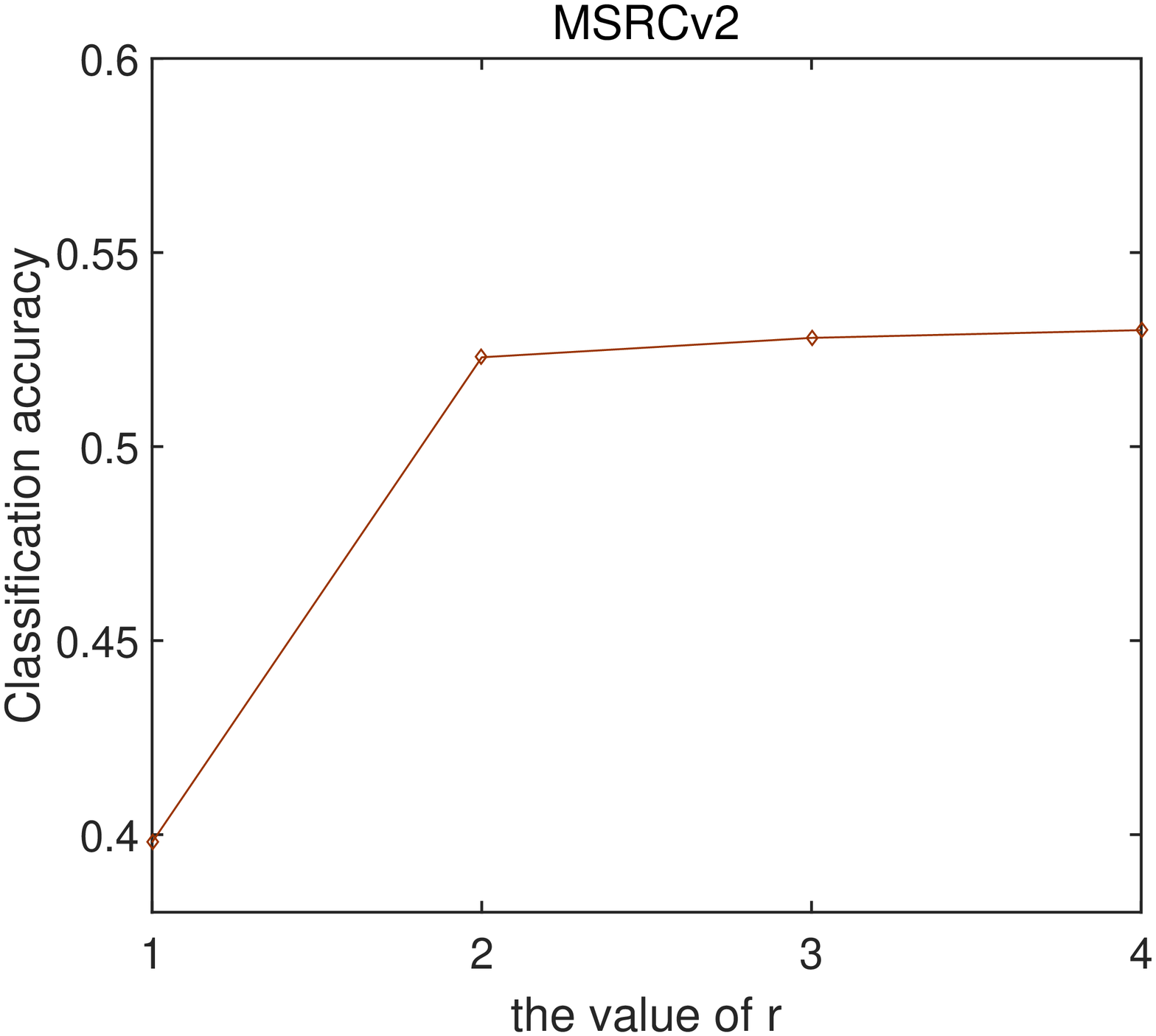}&\includegraphics[width = 2in,height=1.6in]{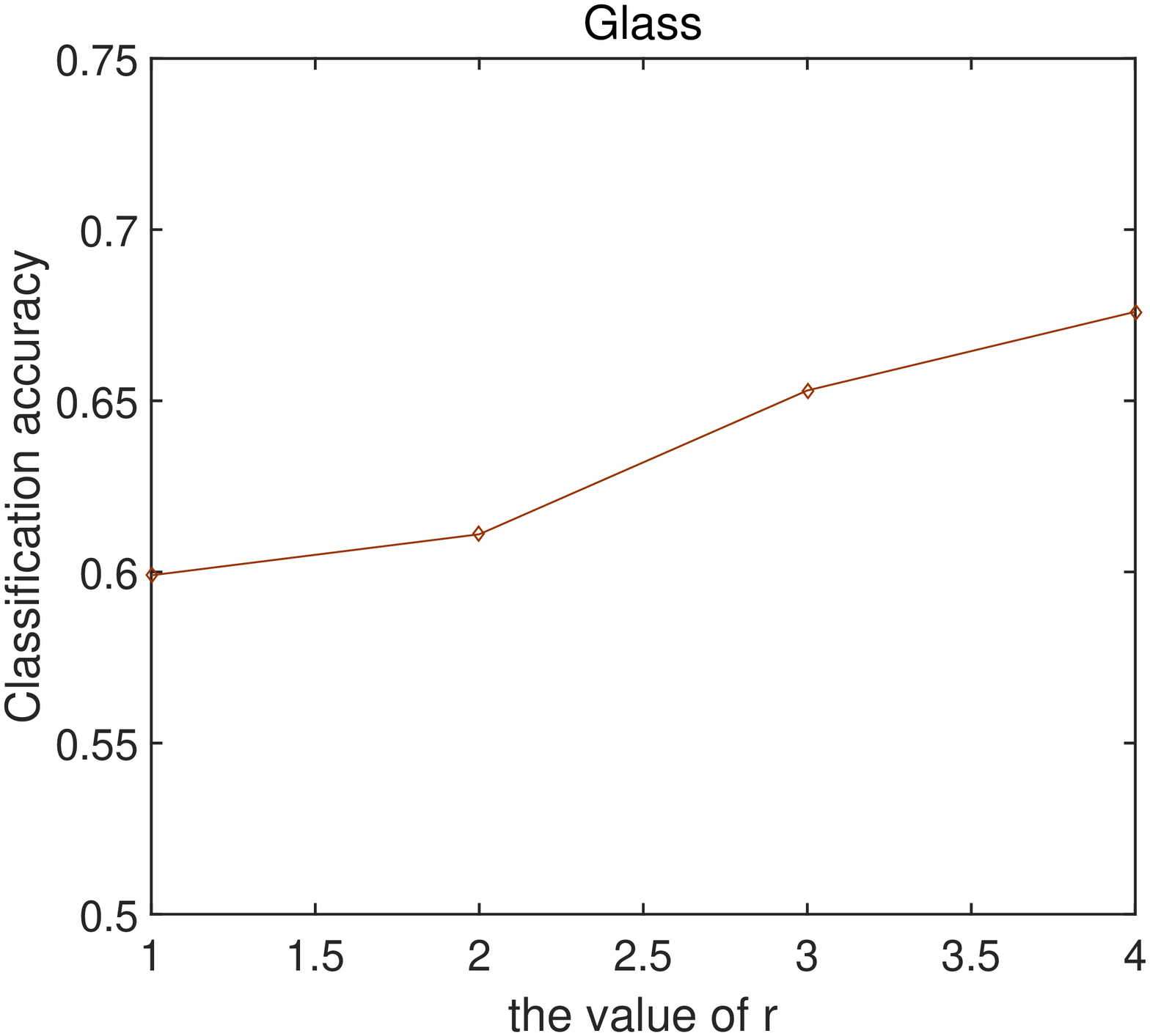}\\
\end{tabular}
\vspace{3mm}
\caption{The classification accuracy of SP-PLL on \emph{Lost}, \emph{MSRCv2} and \emph{Glass} data sets with $\beta$ fixed ($\beta = 0.3$ on \emph{Lost} data set, $\beta = 0.6$ on \emph{MSRCv2} data set and $\beta = 0.7$ on \emph{Glass} data set respectively)}
\label{Fig-analy2}
\vspace{0mm}
\end{figure*}

\subsubsection{\textbf{Real-world (RW) data sets}}
We compare the GM-PLL with all above comparing algorithms on the real-world data sets. The comparison results of inductive accuracy and transductive accuracy are separately reported in Table \ref{Table3} and Table \ref{Tabledisambiguation}, where the recorded results are based on ten-fold cross-validation.

The transductive classification accuracy reflects the disambiguation capacity of PLL methods in recovering ground-truth labeling information from candidate label set, while the inductive classification accuracy reflects the prediction capacity of obtaining the ground-truth label for unseen examples. According to Table \ref{Table3} and Table \ref{Tabledisambiguation}, it is clear to observe that GM-PLL performs better than most comparing PLL algorithms on these RW data sets. The superiority of GM-PLL can be embodied in the following aspects:
\begin{itemize}
\item As shown in Table \ref{Table3}, GM-PLL significantly outperforms all comparing methods on \emph{Lost}, \emph{MSRCv2}, and \emph{SoccerPlayer} data sets, respectively. Especially, compared with the classical methods, the classification accuracy of the proposed method is 40.5\% higher than that of PL-KNN on \emph{Lost} data set, and 20.2\% higher than that of CLPL on \emph{SoccerPlayer} data set. Even compared with the state-of-the-art methods, it also can achieve 2.5\% higher than PL-ECOC on \emph{MSRCv2} and 1.1\% higher than IPAL on \emph{Lost} data set.
\item Meanwhile, GM-PLL also achieves competitive performance on other RW data sets. Specifically, for the \emph{FG-NET} data set, GM-PLL outperforms all comparing methods except PL-LEAF, where it is only 0.7\% lower than PL-LEAF. But on \emph{Yahoo! News} data set, GM-PLL performs great superiority than PL-LEAF, where the classification accuracy is 3.4\% higher than that of PL-LEAF. Besides, among all comparing methods, it is impressive that GM-PLL outperforms CLPL and PL-KNN on all six RW data sets. And, it also exceeds other comparing methods over four in six RW data sets. The experimental results demonstrate the superiority of GM-PLL.
\item As shown in Table \ref{Tabledisambiguation}, GM-PLL shows significantly superior disambiguation ability on \emph{Lost}, \emph{MSRCv2} and \emph{FG-NET} data set and competitive disambiguation ability on \emph{BirdSong} and \emph{SoccerPlayer} data sets, which demonstrates the superiority of the GM scheme on disambiguation. But for \emph{Yahoo! News} data set, GM-PLL is inferior to some comparing state-of-the-art methods. Even so, it can still achieve superior or comparable performance against other comparing methods on making prediction for unseen instances, which demonstrates the superiority of GM scheme on making prediction for unseen instances. In summary, the experimental results demonstrate the effectiveness of our proposed GM-PLL algorithm.
\item We notice that the performance of GM-PLL is inferior to most comparing methods on \emph{Yahoo! News} data set, which is attributed to the low intra-class instance similarity. Especially, over 8440 examples come from two categories, among which the intra-class instance similarity of over 65\% examples is less than 0.60. Obviously, such low intra-class instance similarity may decrease the effectiveness of our proposed method.
\end{itemize}

\subsubsection{Summary}
The two series of experiments mentioned above powerfully demonstrate the effectiveness of GM-PLL, and we attribute the success to the superiority of GM scheme, i.e. simultaneously taking the instance relationship and the co-occurrence possibility of varying instance-label assignments into the same framework. Concretely speaking, for the instance relationship, especially the instance dissimilarity, it can alleviate the effect of the similar instance with varying labels and avoid the outputs of instances be overwhelmed by that of its negative nearest instances. And for the instance-label assignments, the co-occurrence possibility can lead the algorithm to pay more attention to matching selection and reducing its dependence on instance relationship. The two schemes jointly improve the effectiveness and robustness of the proposed method. And as expected, the experimental results demonstrate the effectiveness of our method.

\section{Further Analysis}

\subsection{Parameter Sensitivity}
\label{section-analysis}
The proposed method learns from the PL examples by utilizing two important parameters, i.e. $\beta$ (threshold parameter) and $r$ (the number of candidate labels assigned to unseen instances). Figure \ref{Fig-analy1} and Figure \ref{Fig-analy2} respectively illustrate how GM-PLL performs under different $\beta$ and $r$ configurations. We study the sensitivity analysis of GM-PLL in the following subsection.

\subsubsection{\textbf{The threshold parameter $\beta$}}
The threshold parameter controls the percentage of prior knowledge incorporated into the learning framework. More prior knowledge can be added into the framework as $\beta$ is small, while less prior knowledge contributes to the learning process when $\beta$ becomes larger. On the other hand, small $\beta$ will draw more noise into the learning framework and large $\beta$ will lose more valuable information, two of which have negative effects on the learning model. Faced with varying data sets, we set the threshold parameter $\beta$ among $\{0.3, 0.4, \ldots,0.8\}$ via cross-validation and the specific value is shown in Table \ref{table4}.
\begin{table}[!ht]
\centering
\caption{The optimal value of $\beta$ for GM-PLL}
\vspace{1mm}
\label{table4}
\resizebox{8.5cm}{!}{
\begin{tabular}{c|cccccc}
\cline{1-6}
\hline \hline
Data set     & Lost    & MSRCv2   & FG-NET  &BirdSong     &SoccerPlayer   & Yahoo! News  \\ \hline
$\beta$      & 0.6     & 0.6      &  0.8    &0.5          &  0.3          & 0.7          \\ \hline \hline
\end{tabular}}
\vspace{-1mm}
\end{table}


\subsubsection{\textbf{The number $r$ of candidate label for unseen instances }}
As mentioned above, the percentage of candidate labels assigned to unseen instances has great influence on making prediction for unseen instances. According to the analysis in section \ref{section-prediction}, we simultaneously take the total number of class labels (CL*) and the average number of class labels (AVG-CL*) into consideration, and then utilize Eq (\ref{induce_r}) to obtain the number of assigned labels $r$. To demonstrate the validness of Eq (\ref{induce_r}) empirically, we conduct the experiments under different $r$ configuration and express the comparing results in Figure \ref{Fig-analy2}.

As described in Figure \ref{Fig-analy2}, with the increasing of $r$, the classification accuracy of GM-PLL at first increases and later decreases. And such phenomenon is intuitive, i.e. algorithm with smaller $r$ indicates that less noisy labels need to be removed but the ground-truth label has lower possibility to be contained in the candidate label set; and larger $r$ indicates that the ground-truth label has higher possibility to be contained in the candidate label set but it tends to draw more noisy labels into the candidate label set. The number comparison of assigned candidate labels between empirically optimal value and calculation results of Eq (\ref{induce_r}) on each RW data set is exhibited in Table \ref{table5}. As shown in Table \ref{table5}, except the \emph{FG-NET} data set, the empirically optimal number of candidate labels $r^{**}$ is basically identical to the calculation results $r^{*}$ of Eq (\ref{induce_r}).

\begin{table}[!ht]
\centering
\caption{The number comparison of candidate labels between the optimal value of $r^{**}$ and the calculation results $r^{*}$ of Eq (\ref{induce_r})}
\vspace{1mm}
\label{table5}
\resizebox{8.5cm}{!}{
\begin{tabular}{c|cccccc}
\cline{1-6}
\hline \hline
Data set     & Lost    & MSRCv2   & FG-NET    &BirdSong      &SoccerPlayer   & Yahoo! News  \\ \hline
$r^{*}$      & 3       & 3        &   4       &3             &  2            &  2          \\
$r^{**}$     & 3       & 4        &   1       &4             &  2            &  1           \\ \hline \hline
\end{tabular}}
\vspace{-1mm}
\end{table}

\subsection{Time Consumption}

Although we have conducted corresponding strategies to reduce the computational complexity of the proposed algorithm, the time consumption of the proposed prediction model is still longer than some comparing methods on some large-scale data sets. Nonetheless, such time consumption is acceptable for the PLL problem. Specifically, on most UCI data sets, the time consumptions are no more than 30 seconds; meanwhile, on some small-scale or medium-scale RW data sets, it is also no more than 20 seconds. Moreover, although the time consumption of the prediction model is longer than some comparing methods, the total running time cost (combining training time and testing time) is appropriate and sometimes even less than some state-of-the-art PLL methods, such as PL-LEAF. According to our experimental results, the running time cost of our proposed methods is no more than 1.5h on all RW data sets, which is only 1/10 of that of PL-LEAF. Table \ref{testtimecost} illustrates the total running time and testing time consumption of our proposed algorithm on both UCI and RW data sets, measured within Matlab environment equipped with Intel E5-2650 CPU.

\begin{table}[!ht]
\centering
\caption{Total running time and testing time consumption of our proposed algorithm on UCI and RW data sets}
\vspace{1mm}
\label{testtimecost}
\resizebox{8.5cm}{!}{
\begin{tabular}{c|ccccc}
\cline{1-6}
\hline \hline
Data set           & Lost        & MSRCv2   & FG-NET    &BirdSong      &SoccerPlayer    \\ \hline
running time       & 37.046s     & 127.818s & 198.160s  &281.765s      &3271.877s        \\
testing time       & 0.837s      & 1.431s   & 1.254s    &21.879s       &422.012s        \\ \hline
Data set           & Yahoo! News & glass    & segment   &satimage      &vehicle         \\ \hline
running time       & 8612.220s   & 2.080s   & 80.095s   &236.724s      &7.138s        \\
testing time       & 1025.886s   & 0.204s   & 5.901s    &29.574s       &0.743s          \\ \hline
Data set           & letter      & abalone  & ecoli     &dermatology   &pendigits       \\ \hline
running time       & 312.502s    & 268.547s & 1.916s    &2.924s        &116.202s        \\
testing time       & 28.344s     & 21.380s  & 0.287s    &0.334s        &11.538s         \\ \hline \hline
\end{tabular}}
\vspace{-1mm}
\end{table}


\section{Conclusion}
In this paper, we have proposed a novel graph-matching based partial label learning method GM-PLL. To the best of our knowledge, it is the first time to reformulate the PLL problem into a graph matching structure. By incorporating much prior knowledge and establishing INS-CDL assignments, the proposed GM-PLL algorithm can effectively contribute the valuable information to the learning model. Extensive experiments have demonstrated the effectiveness of our proposed method. In the future, we will further explore other knowledge from PL data and improve the denoising method to further improve the effectiveness and robustness of the model.


\ifCLASSOPTIONcaptionsoff
  \newpage
\fi

%
%
\bibliographystyle{IEEEtran}
\bibliography{sp2017}

\end{document}